\documentclass[10pt,twocolumn,letterpaper]{article}

\usepackage[pagenumbers]{iccv} 

\definecolor{iccvblue}{rgb}{0.21,0.49,0.74}
\usepackage[pagebackref,breaklinks,colorlinks,allcolors=iccvblue]{hyperref}

\usepackage{multirow}
\usepackage{xspace}
\usepackage{listings}
\usepackage{color}
\usepackage{colortbl}
\usepackage{multirow}
\usepackage{amsmath}
\usepackage{hhline}
\usepackage{bigstrut}
\usepackage{rotating}
\usepackage{tcolorbox}
\usepackage{tabularx}
\usepackage{makecell}
\usepackage{booktabs} 
\usepackage{nicefrac}
\usepackage{arydshln} 
\usepackage{wrapfig}

\newcommand{\model}{AdPO}

\newcommand\blfootnote[1]{%
  \begingroup
  \renewcommand\thefootnote{}\footnote{#1}%
  \addtocounter{footnote}{-1}%
  \endgroup
}

\def\paperID{8132} 
\def\confName{ICCV}
\def\confYear{2025}

\title{AdPO: Enhancing the Adversarial Robustness of Large Vision-Language
Models with Preference Optimization}

\author{
Chaohu Liu\textsuperscript{1,2$\S$},
Tianyi Gui\textsuperscript{3}, 
Yu Liu\textsuperscript{3}, 
Linli Xu\textsuperscript{1,2$\dagger$} \\
\textsuperscript{1}School of Computer Science and Technology, 
University of Science and Technology of China\\
\textsuperscript{2}State Key Laboratory of Cognitive Intelligence,
\textsuperscript{3}Tongyi Lab\\
{\tt\small{liuchaohu@mail.ustc.edu.cn, guitianyi@gmail.com, ly103369@alibaba-inc.com}}  \\
{\tt\small{linlixu@ustc.edu.cn}}
}

\begin{document}
\maketitle
\blfootnote{$\S$ Work done during an internship at Tongyi Lab. $\dagger$ Corresponding author. }

\begin{abstract}
Large Vision-Language Models (LVLMs), such as GPT-4o and LLaVA, have recently witnessed remarkable advancements and are increasingly being deployed in real-world applications. 
However, inheriting the sensitivity of visual neural networks, LVLMs remain vulnerable to adversarial attacks, which can result in erroneous or malicious outputs. 
While existing efforts utilize adversarial fine-tuning to enhance robustness, they often suffer from performance degradation on clean inputs. 
In this paper, we proposes \textbf{\model}, a novel adversarial defense strategy for LVLMs based on preference optimization. 
For the first time, we reframe adversarial training as a preference optimization problem, aiming to enhance the model’s preference for generating normal outputs on clean inputs while rejecting the potential misleading outputs for adversarial examples.
Notably, \model~achieves this by solely modifying the image encoder, e.g., CLIP ViT, resulting in superior clean and adversarial performance in a variety of downsream tasks.
Considering that training involves large language models (LLMs), the computational cost increases significantly. 
We validate that training on smaller LVLMs and subsequently transferring to larger models can achieve competitive performance while maintaining efficiency comparable to baseline methods.
Our comprehensive experiments confirm the effectiveness of the proposed \model, which provides a novel perspective for future adversarial defense research.
\end{abstract}
\section{Introduction}
\begin{figure}[t]
\setlength{\abovecaptionskip}{0.cm}
  \centering
  \includegraphics[width=\linewidth]{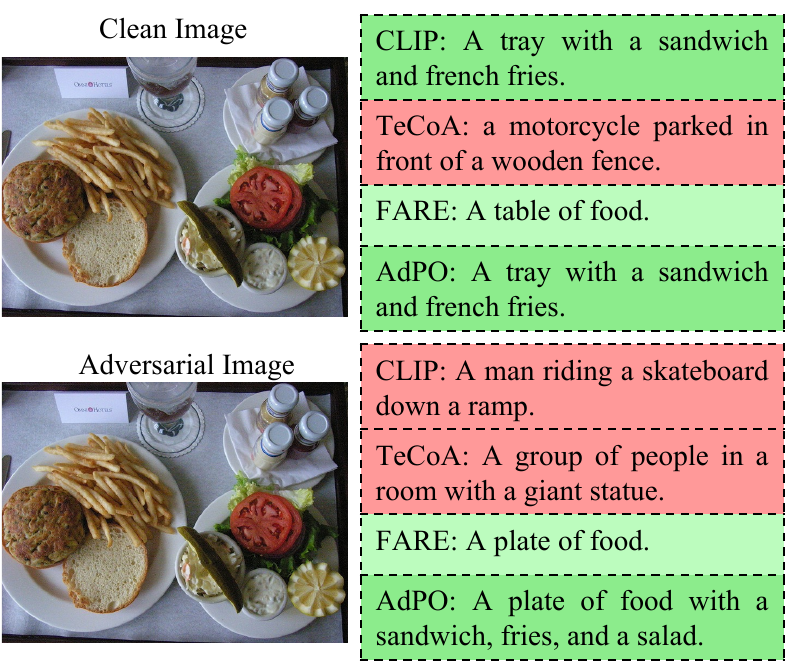 }
  \caption{Illustration of untargeted adversarial attacks on LLaVA using different CLIP models as encoders.
  The original model can produce accurate outputs on clean images, but it makes significant errors when faced with adversarial attacks.
  Although the adversarially trained versions, TeCoA and FARE, have better adversarial robustness, they still tend to hallucinate or fail to fully comprehend the image.
  Comparatively, our \model~exhibits strong performance on both clean and adversarial images.}
  \label{fig:example}
\end{figure}
The emergence of large vision-language models (LVLMs) has substantially propelled the development of general artificial intelligence, attracting considerable attention from the research community~\citep{mllmsurvey2023, survey_mllm_auto, survey_mllm_safety}.
These models generally consist of two key components: visual modules and Large Language Models (LLMs)~\citep{survey_llm}. 
The visual modules, frequently utilizing pre-trained image encoders like CLIP's ViT~\citep{clip}, are responsible for extracting salient visual features from images and projecting them onto the input space of the language model. 
This alignment facilitates the next-token prediction in an autoregressive manner within the framework of the language model.
Cutting-edge LVLMs, such as GPT-4o~\citep{gpt4o}, LLaVA~\citep{llava}, and OpenFlamingo~\citep{openflamingo}, have demonstrated outstanding capabilities in understanding and reasoning with both visual and textual information.
These models have delivered exceptional performance across a broad range of tasks, such as image captioning~\citep{instructblip, improve_mm_image_captioning}, visual question answering~\citep{llava}, and text recognition~\citep{hrvda, monkey, cao2023attention}.

Given their transformative potential in multimodal learning and understanding, LVLMs are increasingly being deployed across a diverse range of real-world applications.
However, this widespread deployment raises significant security concerns, as malicious adversaries can exploit vulnerabilities in LVLMs to induce undesirable outputs and hallucinations~\citep{chiccv2023, jailbreak_pieces}.
Consequently, it is imperative to rigorously test and improve the robustness of these models prior to deployment.
Recent research has identified a critical vulnerability in LVLMs to adversarial attacks targeting both textual and visual inputs~\citep{zhao2023nips}.
Notably, the continuous nature of the visual modality renders it more susceptible to manipulation via numerical optimization techniques~\citep{from_llm_to_mllm_jailbreak, nc2023, visual_jailbreak, CroPA}.
Researchers have demonstrated both targeted and untargeted attacks by introducing imperceptible noise into images, which consequently alters the interpretation and output of the model.

To improve the adversarial robustness of LVLMs, existing efforts  focus on fine-tuning primarily the image encoder~\cite{PMG-AFT, TGA-ZSR, AdvXL}.
For example, TeCoA~\cite{tecoa} uses text-guided contrastive adversarial training to align text and visual features of CLIP, while FARE~\cite{robustclip} introduces unsupervised adversarial fine-tuning to remove the need for labeled datasets at the CLIP embedding level.
Although these methods offer a relatively simple and efficient way to enhance the robustness of CLIP models, they still suffer from significant performance degradation in downstream tasks.
Specifically for LVLMs, such representation-level contrastive learning can lead to a loss of fine-grained image understanding.
As shown in Figure~\ref{fig:example}, TeCoA generates severe hallucinations with clean samples, whereas FARE tends to compromise its ability to capture detailed image features.

Inspired by the significant success of preference optimization in the LLM community~\citep{dpo_survey, rlhf}, we identify that applying preference optimization to adversarial training is highly promising, given the alignment between their objectives.
More specifically, adversarial training aims to enhance model robustness against adversarial attacks while preserving performance on clean data.
Preference optimization, such as DPO~\citep{dpo}, aligns LLMs with human values by increasing the probability of preferred outputs while decreasing the likelihood of non-preferred ones.
Building on this insight,  we propose \textbf{\model}, a novel \textbf{A}dversarial \textbf{d}efense strategy based on \textbf{P}reference \textbf{O}ptimization, which enables LVLMs to generate correct outputs from clean image inputs while rejecting misleading outputs from adversarial images.

However, applying DPO to adversarial training presents unique challenges.
In comparison to standard offline DPO, we introduce two key improvements:
(1) We adapt DPO to an online setting to eliminate the reliance on image annotations, enabling unsupervised training similar to FARE.
The policy model generates interpretations for both clean and adversarial images, which serve as sources for positive and negative samples. 
This online learning strategy offers better data efficiency as it can adapt more quickly to changes in data distribution.
This process is referred to as \textbf{preferred image optimization}.
(2) Multimodal preference optimization may face an \textit{unconditional preference} issue, where the learning process may neglect image conditions~\citep{mdpo}.
To address this issue, we introduce \textbf{adversarial image optimization} to further improve the adversarial robustness.

Another potential concern is computational efficiency.
Directly training a commonly used LVLM model, such as LLaVA-7B~\cite{llava1.5}, may be prohibitively expensive in resource-constrained scenarios.
In this paper, we explore fine-tuning the image encoder of a smaller LVLM and subsequently transferring it to a larger LVLM model.
This strategy not only achieves computational efficiency comparable to previous methods, but also mitigates the risk of potential overfitting during evaluation.

To ensure consistency with previous research~\cite{tecoa, robustclip},  we constrain our adversarial training to modifying only the CLIP ViT parameters on the ImageNet dataset~\citep{imagenet}.
Extensive experimental results, including evaluations on LVLMs and zero-shot classification, demonstrate that our proposed \model~yields a more robust image encoder. 
These findings not only confirm the effectiveness of our approach but also broaden the applicability of preference optimization techniques beyond their traditional use in language models.

In summary, our contributions are as follows:
\begin{itemize}
    \item We introduce \model~(Adversarial defense based on Preference Optimization), which, to the best of our knowledge, is the first attempt to explore the application of preference optimization for adversarial training.
    \item
    We propose the dual strategy of preferred image optimization and adversarial image optimization to maintain the model’s clean performance while enhancing its adversarial robustness.
    \item We validate the feasibility of conducting adversarial training on smaller LVLMs and subsequently transferring it to larger models, which reduces computational costs and mitigates potential overfitting during evaluation.
    \item
    Extensive experiments show that our method achieves state-of-the-art results in improving the adversarial robustness of LVLMs while maintaining the original performance as much as possible.
\end{itemize}

\section{Related Work}
In this section, we primarily review the related studies on large vision-language models, adversarial attacks, adversarial defenses, and preference optimization methods.

\textbf{Large Vision-Language Models.}
Recently, large multimodal models have emerged, including LLaVA 1.5~\citep{llava1.5}, OpenFlamingo (OF)~\citep{openflamingo}, BLIP-2~\citep{blip2}, MiniGPT-4~\citep{minigpt4}, Otter~\citep{otter}, mPLUG-Owl~\citep{mPLUG-Owl}, Qwen-VL~\citep{qwen-vl}, MiniCPM-V~\citep{minicpm-v}, DeepSeek-VL~\citep{deepseek-vl}, InternVL~\citep{internvl2}, and Idefics2~\citep{idefics2}. 
These models typically use pre-trained image encoders (e.g., CLIP or SigCLIP) to extract image features, which are then aligned with text embedding spaces~\citep{clip, sigclip}. 
The visual and textual embeddings are then fed into LLMs for autoregressive generation.
This approach allows the model to simultaneously understand and generate content related to both images and text.
To mitigate computational load, a practical strategy is to freeze the image encoder and train only the projection layer, which not only simplifies the training process but also enhances efficiency~\citep{llava, openflamingo}.
Therefore, image encoders can significantly impact the performance of LVLMs, receiving significant attention from the multimodal community~\citep{SeRum, TinyLLaVA}.
We focus on the performance evaluation of LLaVA-1.5 and OF, as both use CLIP ViT-L/14 ~\citep{clip} as their image encoder.

\textbf{Adversarial attacks.}
The vulnerability of visual neural network models to adversarial attacks is well-established and has been extensively investigated~\citep{adv_example, fsgm, pgd, adv_patch, tgr, zhang2024adv10year, advclip, wang2025tracking}.
By introducing carefully crafted noise into images, adversaries can cause the victim model to generate incorrect outputs with high confidence.
Capitalizing on this vulnerability, recent studies have shown that LVLMs are also vulnerable to attacks targeting visual inputs~\cite{chiccv2023, jailbreak_pieces, CroPA, verbose_image, bard_robust}.
Zhao \textit{et al.}~\cite{zhao2023nips} showed that transferable black-box attacks could be generated using text-to-image models and other work~\cite{nc2023} demonstrated how adding adversarial noise to images can circumvent safety constraints of LLMs. 
Qi \textit{et al.}~\cite{qi2024aaai} explored how adversarial attacks embedding deceptive information into images can mislead LVLMs and deceive users.
The widespread deployment of LVLMs has raised urgent security concerns due to the threat of adversarial attacks.

\textbf{Adversarial defenses.}
Adversarial defenses in machine learning safeguard models from malicious inputs to ensure their integrity and reliability, especially in security-sensitive contexts~\citep{pgd, MirrorCheck, distillation_adv, deep_metric_learning, luo2024game, ledda2024advtrain, debbi2024CausAdv, xue2024advtrain, zhao2024advtrainsurvey, liang2024advdefensevlm, li2024dat, li2024advtrain, hotegni2024advtrain, jiang2024advtrain}. 
For example, Detectors~\citep{huang2024advdetection, mumcu2024detect, mavali2024detect, kevin2019detect, weilin2018detect, MagNet, jan2017detect} identify and filter out adversarial examples, but these external modules can introduce additional inference time and may also obstruct normal inputs.
Purification methods~\cite{pouya2018defense, weili2022defense, chin2022defense, nilaksh2018defense} use techniques such as diffusion models to eliminate adversarial perturbations in input data, and this can also modify the input, thus affecting performance.
Adversarial training~\cite{adml, ensembleadvtrain, uiat, trap-mid, xiaojun2024pgk, lv2024advtuning, depalma2024advtrain, dong2024advtrain, ribeiro2024advtrain, xiaojun2022lasat} is a foundational method for enhancing a model's inherent robustness by integrating adversarial examples into the training dataset.
In the multimodal domain~\cite{PMG-AFT, TGA-ZSR, AdvXL}, TeCoA improves the adversarial robustness of CLIP's image encoder through text-guided contrastive adversarial training while preserving some of CLIP's zero-shot classification capabilities~\citep{tecoa}. 
FARE employs unsupervised training by minimizing the distance between adversarial image features and clean image features, maintaining impressive performance on LVLMs~\citep{robustclip}. 
However, this straightforward adversarial training approach often fails to prevent performance degradation on clean samples. 
Unlike these fine-tuning strategies, we are the first to frame adversarial training as a preference optimization problem, integrating both clean and adversarial images into the training process to improve robustness while maintaining clean performance.

\textbf{Preference optimization.}
Preference optimization has emerged as a novel training paradigm for aligning LLMs with human values and has garnered significant attention in recent research~\citep{clipdpo, rlhfv, rlaifv, mdpo, sima}. 
Reinforcement Learning from Human Feedback (RLHF) utilizes human preferences as a reward model and applies reinforcement learning to guide model training~\citep{rlhf0, rlhf}
Direct Preference Optimization (DPO) streamlines the training process by increasing the log probability of preferred samples while reducing that of non-preferred samples, enabling broader applications~\citep{dpo}.
Subsequent advancements, such as StepDPO~\citep{stepdpo}, SimPO~\citep{simPO}, and IPO~\citep{ipo}, have further improved DPO’s performance.
Considering its stability and efficiency in training, we also adopt DPO for adversarial training of LVLMs in this work.

\section{Method}
\begin{figure}[t]
  \centering
  \setlength{\abovecaptionskip}{0.cm}
  \includegraphics[width=\linewidth]{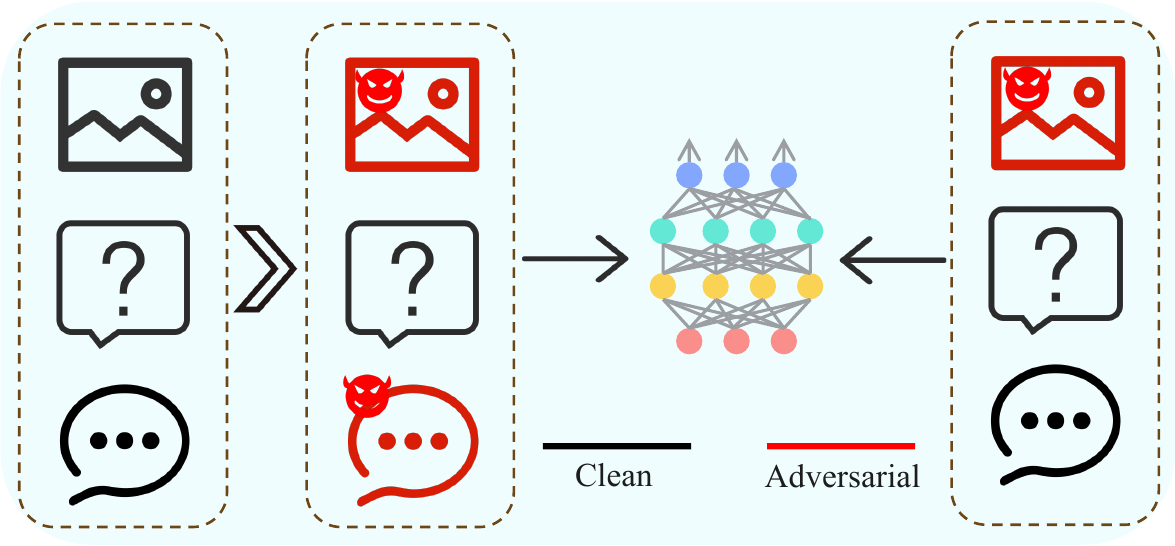 }
  \caption{The architecture of our proposed \model.
  AdPO mainly consists of two parts: (\textbf{left}) preferred image optimization and (\textbf{right}) adversarial image optimization.
  Preferred image optimization incorporates both clean and adversarial images into adversarial training while maintaining the model’s performance on clean inputs, and adversarial image optimization can significantly enhance the model’s adversarial robustness.
  }
  \label{fig:overview}
\end{figure}

This section provides a detailed introduction to our \model, with its overall framework illustrated in Figure~\ref{fig:overview}.
First, Section~\ref{sec:pre} outlines the basics of the DPO algorithm, and Section~\ref{sec:adv_img} discusses adversarial example generation, which forms the preference sample pairs required for DPO.
Sections~\ref{sec:pio} and ~\ref{sec:aio} introduce preferred image optimization and adversarial image optimization, respectively.

\subsection{Preliminaries}
\label{sec:pre}

DPO has emerged as a prominent method in the domain of offline preference optimization.
This method provides a novel framework for optimizing language models in accordance with human preferences.
In a typical setup, given an input $x$ and an output text $y$, a language model (i.e., policy model) $\pi_\theta$ generates a conditional distribution $\pi_\theta(y|x)$. 
Unlike RLHF, which employs an explicit reward model, DPO reformulates the reward function  using a closed-form expression with respect to the optimal policy.
The main objective of DPO is to maximize the expected reward of the outputs generated by this policy, with the reward function defined as $r(x, y)$:
\begin{equation}
r(x,y) = \beta \log \frac{\pi_\theta(y|x)}{\pi_\text{ref}(y|x)} + \beta \log Z(x),
\end{equation}
where $\beta$ is a constant, $\pi_{\text{ref}}$ is the reference policy model (identical to the original $\pi_\theta$), and $Z(x)$ is the partition function.

Given a preference dataset $\mathcal{D} = \{x, y_w, y_l\}$, where $y_w$ and $y_l$ represent the winning and losing responses respectively, DPO employs a Bradley-Terry model \citep{bt} to express the probability for each preference pair:
\begin{equation}
p(y_w \succ y_l) = \sigma(r(x,y_w) - r(x,y_l)),
\end{equation}
where $\sigma(\cdot)$ is typically defined as a sigmoid function.
The key innovation of DPO is its formulation of the likelihood of preference data using the policy model, as opposed to relying on an explicit reward model.
This leads to the formulation of the DPO objective:
\begin{equation}
\begin{aligned}
&\mathcal{L}_{\mathrm{DPO}}(\pi_\theta;\pi_{\mathrm{ref}}) =-\mathbb{E}_{(x,y_w,y_l)\sim\mathcal{D}} \\
&\left[\log\sigma\left(\beta\log\frac{\pi_\theta(y_w|x)}{\pi_{\mathrm{ref}}(y_w|x)} - \beta\log\frac{\pi_\theta(y_l|x)}{\pi_{\mathrm{ref}}(y_l|x)}\right)\right],
\end{aligned}
\end{equation}
This formulation captures the core principles of DPO, providing a robust framework for optimizing language models in alignment with human preferences.

\subsection{Adversarial Example Generation}
\label{sec:adv_img}

In the context of LVLMs, the input to the model comprises \( x = \{x_{m}, x_{text}\} \), where \( x_{m} \) denotes the image input and \( x_{text} \) represents the text input. 
This section outlines the principles behind generating adversarial images.

Adversarial images are generated by introducing small, nearly imperceptible perturbations to original images, with the goal of deceiving machine learning models and inducing incorrect predictions~\citep{adv_example, fsgm}.
Although adversarial images appear nearly identical to the original images to humans, they can drastically alter the model’s output, exposing its vulnerability to malicious inputs~\citep{ad_ex_physicial_world}.
Adversarial attacks can be broadly categorized into targeted and untargeted attacks: targeted attacks compel the model to produce specific outputs~\citep{CroPA}, whereas untargeted attacks merely lead the model to generate incorrect outputs~\citep{wang2024break, verbose_image}.
In this study, we employ untargeted attack methods to generate adversarial images for three reasons:
(1) They eliminate dependence on specifically labeled datasets and do not rely on the text encoder, enabling our method to generalize to unseen datasets~\cite{robustclip}.
(2) Untargeted attacks typically achieve a high success rate, allowing the stable generation of  negative adversarial samples during training~\cite{cui2024cvpr}.
(3) Their broader attack capability enhances the model’s resilience against various types of adversarial attack methods~\cite{wang2024break}.

Given an image encoder $\phi$ (e.g., CLIP ViT) and a clean image $x_{m}$, adversarial examples are generated by optimizing to maximize the discrepancy between the encoded features of the adversarial image and the clean image:
\begin{equation}
x_{adv} = \mathop{\arg\max}_{\left\|x_{adv}- x_{m}\right\|_\infty\leq\varepsilon}\left\|\phi(x_{adv})-\phi_{\mathrm{org}}(x_{m})\right\|_2^2,
\label{adv_image}
\end{equation}
where $x_{adv}$ is the adversarial image obtained through iterative optimization like PGD~\citep{pgd}, $\phi_{org}$ is the original image encoder and $\epsilon$ is the image perturbation magnitude.
Note that in subsequent adversarial training, the parameters of $\phi$ will be updated.

\subsection{Preferred Image Optimization}
\label{sec:pio}

This section primarily delineates the methodology for constructing pairs of preferred and non-preferred samples from unlabeled image data, a fundamental step in the DPO training pipeline.

\textbf{Model Selection.}
Compared to previous methods~\cite{tecoa, TGA-ZSR, PMG-AFT} that rely solely on CLIP’s image and text encoders, \model~utilizes the entire LVLM model. 
Using a commonly adopted model such as LLaVA-7B~\cite{llava1.5} would result in high computational costs. 
To address this, we construct TinyLLaVA\footnote{\text{https://github.com/TinyLLaVA/TinyLLaVA\_Factory}}, which leverages OpenELM-450M-Instruct~\cite{Openelm} as its language model. This lightweight LVLM not only achieves training efficiency comparable to previous approaches but also mitigates potential overfitting during evaluation.

Given a clean image $x_{m}$ and its adversarial image $x_{adv}$, we employ an online approach to directly prompt the model (e.g., \textit{“What is the content of the image?”}) to generate interpretations, thereby obtaining the preferred response $y_w$ and the non-preferred response $y_l$.
Accordingly, in the setting of multimodal adversarial training, our preferred image optimization can be formulated as:
\begin{equation}
\begin{aligned}
&\mathcal{L}_{\mathrm{P}}(\pi_\theta;\pi_{\mathrm{ref}}) \\
&= -\mathbb{E}_{(x_{m}, x_{text}, y_w, y_l) \sim \mathcal{D}} \\
&\quad \log\sigma\Biggl( \beta \log \frac{\pi_\theta(y_w \mid x_m, x_{text})}{\pi_{\mathrm{ref}}(y_w \mid x_m, x_{text})} \\
&\qquad - \beta \log \frac{\pi_\theta(y_l \mid x_{adv}, x_{text})}{\pi_{\mathrm{ref}}(y_l \mid x_{adv}, x_{text})} \Biggr),
\end{aligned}
\end{equation}
This straightforward approach presents several advantages.
First, it removes the need for data annotation, thus facilitating its application to previously unseen image data.
Second, this method resembles semi-supervised learning, especially as LVLMs now possess advanced capabilities, enabling them to incorporate labeled images into their training data. 
Moreover, allowing the model to generate its own text as labels effectively mitigates distribution shift issues, thus concentrating attention on the adversarial images themselves~\citep{silkie}.

Given the rapid development of preference optimization algorithms, we will evaluate the performance of DPO variants in experiments to assess the adaptability of \model.

\subsection{Adversarial Image Optimization}
\label{sec:aio}
Although preferred image optimization can maintain the performance of LVLMs on clean inputs, it does not significantly enhance adversarial robustness.
Recent research indicates that, although multimodal DPO is designed to compute implicit rewards based on all input modalities, it can prioritize language-only preferences while neglecting image conditions (i.e., unconditional preferences), resulting in suboptimal model performance and increased hallucinations~\citep{mdpo}.

The issue of unconditional preferences may lead to suboptimal adversarial robustness. To address this, we introduce simple yet effective adversarial image optimization:
\begin{equation}
\mathcal{L_{\mathrm{A}}}=-\sum_{t=1}^{T}\log \pi_{\theta}(y_w^{t}\mid x_{adv},x_{text},y_{w}^{0:t-1}),
\end{equation}
where $T$ represents the token length of $y_w$. 
Notably, the language modeling loss is calculated exclusively over the answer portion of the sequence rather than the entire input-output chain. This targeted loss computation explicitly focuses the model's optimization effort on semantic fidelity of critical response segments, while mitigating overfitting to non-essential contextual tokens.

The objective of \model~is a combination of  preferred image optimization and adversarial image optimization:
\begin{equation}
    \mathcal{L}_\mathrm{\model} = \mathcal{L_{\mathrm{P}}} + \lambda \mathcal{L_{\mathrm{A}}}. 
\end{equation}
where $\lambda$ is the scaling factor that balances the two loss terms. Empirically, we find that increasing $\lambda$ enhances adversarial performance, while decreasing it improves clean performance.

By leveraging joint optimization, \model~attains enhanced adversarial robustness while maintaining its performance on clean samples.
\section{Experiments}
In this section, we conduct extensive experiments to evaluate the performance of \model~on LVLMs, with additional zero-shot classification experiments provided in the appendix.
Although we employ full LVLMs during adversarial training, we modify only the parameters of their image encoder. This allows the robust image encoder to be directly transferred to other LVLMs while preserving the capabilities of the language model.

\textbf{Models.}
For training, we use TinyLLaVA~\cite{jia2024tinyllava}, which consists of CLIP’s ViT-L/14~\citep{clip} as the image encoder and OpenELM-450M-Instruct as the language model. This lightweight architecture ensures computational efficiency comparable to previous methods while mitigating potential overfitting during subsequent evaluations.
For evaluation, we primarily select OpenFlamingo-9B (OF)\citep{openflamingo} and LLaVA 1.5-7B\citep{llava1.5}, both of which use the same image encoder.
The two models differ in their language decoders: OF employs MPT-7B~\citep{mpt7b}, while LLaVA 1.5 uses Vicuna~\citep{vicuna}.
In the subsequent evaluation of OF, we adopt a zero-shot setting, where the model is given textual prompts without any accompanying images~\citep{flamingo}.
For LLaVA, we employ the default system prompt along with task-specific prompts~\citep{llava}.

\textbf{Adversarial training settings.}
For a fair comparison, we conduct training on the ImageNet dataset~\cite{imagenet}.
As we adopt an online learning approach, we do not rely on category labels provided by the dataset, only on the images themselves.
By optimizing Equation~\ref{adv_image}, we generate adversarial perturbations for clean images using a 10-step PGD under the $\ell_{\infty}$ norm.
It is widely recognized that employing larger image perturbations during adversarial training can significantly improve adversarial robustness, but it often leads to performance degradation on clean data~\citep{pgd}.
To balance robustness and clean accuracy, we apply two perturbation radii: $\epsilon=\nicefrac{2}{255}$ and $\epsilon=\nicefrac{4}{255}$.
The resulting robust CLIP image encoders are referred as \model$^2$ and \model$^4$, respectively. 
$\lambda$ is set to 1 by default.
We use the AdamW optimizer with a weight decay of 1e-4 and a learning rate of 1e-5.
We conduct training for two epochs with a batch size of 128. 
The preference  parameter $\beta$ is set to 0.1.

\textbf{Baseline methods.}
We compare the performance of AdPO with the original CLIP and state-of-the-art methods, TeCoA~\citep{tecoa}, FARE~\citep{robustclip}, and TGA-ZSR~\cite{TGA-ZSR}.
TeCoA and TGA-ZSR utilize supervised contrastive learning with image category labels, while FARE performs unsupervised training at the representation level.
To ensure fair comparison, we use adversarial images with the same noise radius for training, denoted as TeCoA$^2$, TGA$^2$, and FARE$^2$ for $\epsilon=\nicefrac{2}{255}$, and TeCoA$^4$, TGA$^4$, and FARE$^4$ for $\epsilon=\nicefrac{4}{255}$.

\subsection{Evaluation of Untargeted Attacks on LVLMs}
\label{sec:eval_untargeted_attack}
In this section, we evaluate the clean and adversarial  performance of \model~in vision-language tasks by replacing the image encoder of LVLMs with robust versions.

\textbf{Attack setup.}
We utilize the approach outlined in \cite{chiccv2023} to perform untargeted attacks aimed at degrading the model’s performance.
Given that attacks on LVLMs often demand more iterations, we employ a 100-step APGD attack~\citep{apgd}, which utilizes ground-truth captions as labels.
After each attack, we discard samples with scores below a specified threshold to ensure that computationally expensive attacks are only performed when necessary, following \cite{robustclip}.
Further details are provided in the Appendix.

\begin{table*}[t!]
\centering
\setlength{\abovecaptionskip}{0.cm}
\setlength{\tabcolsep}{2.3mm}{
\begin{tabular}{ll||ccc|ccc|ccc|ccc} 
\toprule
 \multirow{3}{*}{VLM} &  \multirow{3}{*}{ \makecell{Image \\ Encoder} } & \multicolumn{3}{c}{COCO}  &  \multicolumn{3}{c}{Flickr30k} &  \multicolumn{3}{c}{TextVQA} & \multicolumn{3}{c}{VQAv2} \\
 \cline{3-14}
 &   &  \multirow{2}{*}{clean}  & \multicolumn{2}{c|}{$\ell_{\infty}$}  & \multirow{2}{*}{clean}  & \multicolumn{2}{c|}{$\ell_{\infty}$}  & \multirow{2}{*}{clean} & \multicolumn{2}{c|}{$\ell_{\infty}$} & \multirow{2}{*}{clean} & \multicolumn{2}{c}{$\ell_{\infty}$} \\
\cline{4-5}
\cline{7-8}
\cline{10-11}
\cline{13-14}

 &  &  & $\nicefrac{2}{255}$  & $\nicefrac{4}{255}$ &  & $\nicefrac{2}{255}$ & $\nicefrac{4}{255}$ & & $\nicefrac{2}{255}$ &  $\nicefrac{4}{255}$ &  & $\nicefrac{2}{255}$ & $\nicefrac{4}{255}$ \\ \toprule

\multirow{9}{*}{\begin{sideways}\textbf{OF-9B}\end{sideways}} & CLIP & 79.7 & 1.5 & 1.1 & 60.1 & 0.7 & 0.4 & 23.8 & 0.0 & 0.0 & 48.5 & 1.8 & 0.0\\
\cdashline{2-14}
 & TeCoA$^2$ & 73.5 & 31.5 & 21.2 & 49.5 & 14.1 & 9.5 & 16.6 & 3.5 & 2.1 & 46.2 & 23.5 & 20.5 \\
  & TGA$^2$ & 74.1 & 34.3 &  22.5 & 51.9 &  17.1 & 10.2 & 18.2 &  4.2 & 2.1 & 45.3 & 24.7 & 16.5 \\
  & FARE$^2$ & 79.1 & 34.2 & 19.5 & 57.7 & 16.4 & 8.9 & 21.6 & 4.1 & 1.9 & 47.0 & 24.0 & 17.2 \\
 & \model$^2$ & \textbf{79.6} & \textbf{38.2} & \textbf{29.4} & \textbf{60.1} &  \textbf{20.5} & \textbf{14.7} & \textbf{22.5} & \textbf{9.5} & \textbf{4.2} & \textbf{48.2} & \textbf{28.1} & \textbf{23.2}\\

\cdashline{2-14}

 & TeCoA$^4$ & 66.9 & 28.5 & 21.6 & 40.9 & 12.0 & 10.3 & 15.4 & 2.1 & 1.8 & 44.8 & 23.6 & 21.3 \\
& TGA$^4$ & 67.2 & 31.2 & 22.4 & 41.4 &  15.9 & 12.3 & 16.2 & 4.4 & 3.1 & 44.9 & 23.4  & 21.5  \\
& FARE$^4$ & 74.1 & 30.9 & 22.8 & 51.4 & 15.7 & 10.5 & 18.6 & 3.4 & 2.9 & 46.1 & 23.6 & 21.0 \\
 & \model$^4$ & \textbf{77.1}  &  \textbf{36.7} & \textbf{25.9}  & \textbf{57.5} &  \textbf{18.4} & \textbf{12.9} & \textbf{21.9} & \textbf{6.8} & \textbf{3.9} & \textbf{47.5} & \textbf{25.9} & \textbf{21.9}  \\ 
\hline

\multirow{9}{*}{\begin{sideways}\textbf{LLaVA 1.5-7B}\end{sideways}} & CLIP & 115.5 & 4.0 & 3.1 & 77.5 & 1.6 & 1.0 & 37.1 & 0.5 & 0.0 & 74.5 & 2.9 & 0.0\\
\cdashline{2-14}
& TeCoA$^2$ & 98.4 &  44.2 & 30.3 & 57.1 & 23.2 & 15.3 & 24.1 & 12.1 & 8.8 & 66.9 & 33.8 & 21.8  \\
& TGA$^2$  & 108.5 & 55.6 & 31.1 & 68.3 & 28.6 & 17.7 & 28.9  & 14.5 & 8.7 & 70.9 & \textbf{35.1} & 23.1  \\

& FARE$^2$ & 109.9 & 53.6 & 31.0 & 71.1 & 29.5 & 17.5 & 31.9 & 14.7 & 9.1 & 71.7 & 34.9 & 23.0\\

& \model$^2$ & \textbf{114.3} & \textbf{62.7} & \textbf{40.7} &  \textbf{73.8}  &  \textbf{32.4} & \textbf{18.1} &  \textbf{32.2} & \textbf{16.5} & \textbf{10.2} & \textbf{71.9} & 34.1 & \textbf{23.1} \\

\cdashline{2-14}

& TeCoA$^4$ & 88.3 & 50.9 & 35.3 & 48.6 & 27.9 & 19.5 & 20.7 & 12.6 & 9.3 & 63.2 & 41.0 & 31.7  \\
 & TGA$^4$ & 100.2 & 58.2 & 41.8 & 58.2 & 30.5 & 21.9 & 23.8 &  15.8 & 10.8 & 63.1 & 40.9 & 31.5 \\
& FARE$^4$ & 102.4 & 57.1 & 40.9 & 61.6 &  31.4 & 22.8 &  27.6 & 15.8 & \textbf{10.9} & 68.3 & 40.7 & 30.5\\
& \model$^4$ & \textbf{109.2} & \textbf{65.3} & \textbf{47.8} & \textbf{65.7} & \textbf{33.8} & \textbf{24.1} & \textbf{30.0} & \textbf{16.0} & 10.1 & \textbf{70.0} & \textbf{41.1} & \textbf{32.4}\\
\bottomrule
\end{tabular}}
\caption{
Evaluation of the adversarial robustness of large vision-language models with different CLIP models.
We evaluate the clean performance and adversarial robustness of various methods across multiple tasks and perturbation sizes. The results indicate that \model~significantly exceeds our baseline methods, attaining outstanding robustness along with exceptional clean performance. The best results are shown in \textbf{bold}.
} 
\vspace{-0.4cm}
\label{tab:main_vlm}
\end{table*}

\textbf{Datasets and metrics.}
We utilize a variety of datasets for image captioning tasks, including COCO~\citep{coco} and Flickr30k~\citep{flickr30k}, as well as for visual question answering tasks, such as VQAv2~\citep{vqav2} and TextVQA~\citep{textvqa}.
Considering that adversarial attacks are time-consuming and costly, we randomly selected 500 images for evaluation.
We employ the CIDEr score~\citep{cider} for image captioning and VQA accuracy~\citep{vqa} for visual question answering tasks to present our results.

Table~\ref{tab:main_vlm} summarizes the experimental results.
Typically, the original CLIP model achieves optimal clean performance but lacks adversarial robustness, rendering it vulnerable to adversarial attacks.
Our \model~consistently achieves superior clean performance and adversarial robustness compared to baseline methods, emphasizing the significance of including both clean and adversarial images in the training dataset.
Notably, the performance improvements observed on both the OF and LLaVA indicate that the learned features exhibit strong generalization capabilities. This is particularly significant for the OF model, which differs substantially in architecture from the TinyLLaVA used during training.
This also demonstrates the feasibility of training on smaller models and transferring to larger ones, which has significant implications for enhancing the robustness of larger models.
However, previous adversarial training methods, while efficient in leveraging CLIP alone, fail to align well with LVLM tasks, resulting in more significant performance degradation.
For different perturbation sizes, $\epsilon = \nicefrac{2}{255}$ already ensures solid adversarial robustness, while larger perturbations still preserve more clean performance.
AdPO$^4$ exhibits stronger robustness compared to \model$^2$, but at the cost of some clean performance.

\subsection{Evaluation of Targeted Attacks on LVLMs}
\label{sec:target}
\begin{table*}
    \centering
    \setlength{\abovecaptionskip}{0.cm}
    \setlength{\tabcolsep}{0.7mm}{
    \begin{tabular}{l|ccccccccc}
        \toprule
        Target & CLIP & TeCoA$^2$  & TGA$^2$ & FARE$^2$ & \model$^2$ & TeCoA$^4$  & TGA$^4$ &  FARE$^4$  & \model$^4$ \\
        \midrule
        \texttt{A group of people are playing}\dots & 20\,/\,20 & 1\,/\,20 &  2\,/\,20 & 1\,/\,20   & 0\,/\,20 & 0\,/\,20 & 0\,/\,20  & 0\,/\,20 & 0\,/\,20 \\
        \texttt{A group of people are flying} \dots & 20\,/\,20 & 1\,/\,20 & 1\,/\,20  & 1\,/\,20  & 0\,/\,20 & 0\,/\,20 & 0\,/\,20  & 0\,/\,20 & 0\,/\,20 \\
        \texttt{The pizza on the table}\dots & 20\,/\,20 & 2\,/\,20  & 0\,/\,20  &  0\,/\,20 & 0\,/\,20 & 0\,/\,20  & 0\,/\,20 & 0\,/\,20 & 0\,/\,20 \\
        \texttt{An earthquake is about}\dots & 20\,/\,20 & 2\,/\,20 & 1\,/\,20  & 1\,/\,20  & 0\,/\,20  & 0\,/\,20 & 0\,/\,20 & 0\,/\,20 & 0\,/\,20 \\
        \texttt{This patient needs the best}\dots & 20\,/\,20 & 0\,/\,20 & 1\,/\,20 &  0\,/\,20 &   0\,/\,20 & 0\,/\,20 & 0\,/\,20 & 0\,/\,20 & 0\,/\,20 \\
        \midrule
        \textbf{Mean ASR:} & 100\% & 4\%  & 5\%  & 3\% &   \textbf{0\%} &\textbf{ 0\%} & \textbf{0}\%  &  \textbf{0}\% & \textbf{0\%}\\
        \bottomrule
    \end{tabular}}
    \caption{Quantitative evaluation of  targeted attacks at $\epsilon=\nicefrac{4}{255}$ radii. We assess the Attack Success Rate (ASR) for each setup. 
    }
    \vspace{-0.4cm}
\label{tab:targetattack}
\end{table*}
In contrast to the untargeted attacks discussed in Section~\ref{sec:eval_untargeted_attack}, targeted attacks on LVLMs pose a significantly greater threat.
Targeted attacks aim to compel the model to produce specific outputs, with the added noise in the image remaining imperceptible to the user.
Through image manipulation, attackers can circumvent the model’s security mechanisms, leading it to generate malicious content~\citep{nc2023, jailbreaking_vlms, visual_jailbreak}.
Additionally, attackers can embed phishing links into images through adversarial attacks to deceive users~\citep{absuing}.
In this section, we examine the robustness of substituting the CLIP encoder in LLaVA with our adversarially robust variant.

\textbf{Attack setup.}
We perform targeted attack experiments on LLaVA 1.5-7B, using the attack success rate (ASR) as the primary evaluation metric.
A sample is deemed successfully attacked if the model’s output contains the target string.
Targeted attacks on LVLMs generally require more iterations, prompting us to execute APGD attacks for 10,000 iterations.
Given that larger image perturbations pose more significant threats, we employ $\ell_{\infty}$ threat models with a radius of $\epsilon = \nicefrac{4}{255}$.
We evaluate five target strings incorporating errors such
as incorrect medical diagnoses and fake news, sampling 20 images for each string.

The quantitative evaluation results are presented in Table~\ref{tab:targetattack}. The attack success rate for the clean version of the CLIP model reaches 100\%, underscoring the vulnerability of current vision-language models to visual input and the substantial security risks posed. 
TeCoA$^2$, FARE$^2$, TGA$^2$, and \model$^2$ demonstrate varying degrees of adversarial robustness, even when subjected to higher levels of adversarial noise. 
By comparison, the $\epsilon=\nicefrac{4}{255}$ versions exhibit significantly higher levels of adversarial robustness.
Additional details are provided in Appendix.

\begin{figure*}[t]
  \centering
  \setlength{\abovecaptionskip}{0.cm}
  \includegraphics[width=\linewidth]{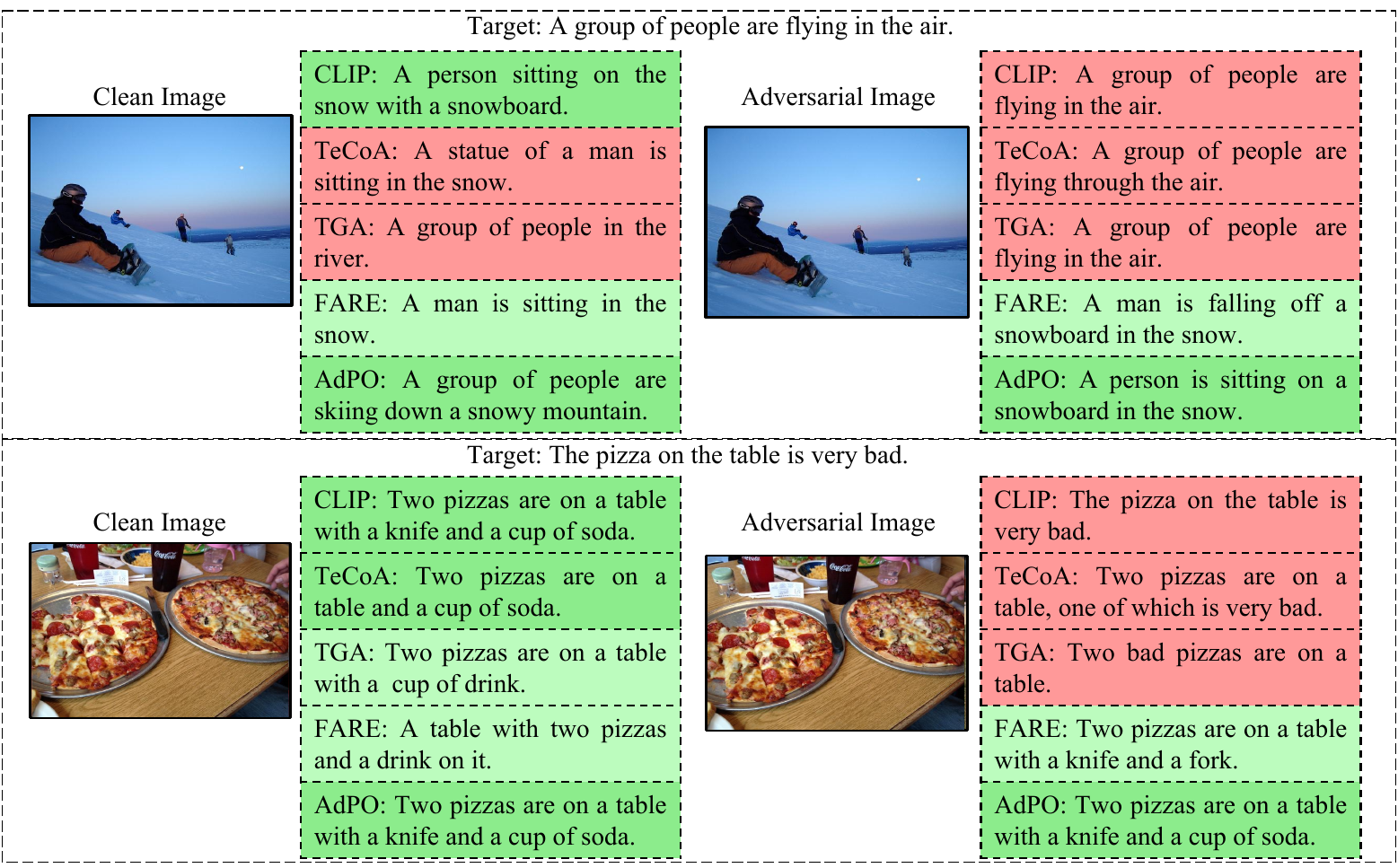 }
  \caption{Qualitative assessment of targeted attacks 
  on LLaVA.
  (\textbf{Left}) When encountering clean images, CoTeA may exhibit noticeable errors, which is undesirable in adversarial defense, while FARE and AdPO demonstrate better clean performance. 
  (\textbf{Right}) When faced with adversarial images, the original CLIP version of LLaVA is easily compromised, FARE shows some adversarial robustness but loses more details or makes subtle errors, whereas AdPO performs better.}
  \vspace{-0.6cm}
  \label{fig:case}
\end{figure*}

\subsection{Further Evaluation}
Although we conduct extensive quantitative evaluations above, they are still insufficient for a comprehensive assessment of LVLMs.
In this section, we first present a qualitative evaluation, followed by an analysis of other vision-language tasks and the training efficiency.

\textbf{Qualitative evaluation.} As depicted in Figure~\ref{fig:case}, the LLaVA model, using the original CLIP as the encoder, provides the most accurate and detailed understanding of clean images.
However, when faced with adversarial images generated by targeted attacks, they are completely vulnerable to successful attacks.
TeCoA fails to exhibit robust performance against both clean and adversarial images, whereas FARE experiences a loss of detail or minor errors in image understanding, ultimately falling short of optimal performance.
In the absence of adversarial defenses, LLaVA is susceptible to manipulation, resulting in biased outputs that can mislead users and have detrimental effects.
Therefore, it is imperative to enhance the model’s adversarial robustness.

Recent work has shown that LVLMs are prone to hallucinations and are more susceptible to jailbreak attacks compared to purely language models~\cite{qi2024aaai, pope}.
Additional experimental evaluations presented in Appendix demonstrate that our method exhibits better performance in both hallucination reduction and jailbreak prevention.
We also find that \model demonstrates higher data utilization efficiency, achieving competitive results with only half of the training data. 
For more details, please refer to Appendix.

\subsection{Ablation Study}

        

In this section, we conduct ablation studies to analyze the factors influencing \model.

\begin{table}[t]
\setlength{\abovecaptionskip}{0cm}
\centering
\setlength{\tabcolsep}{0.3mm}{
\begin{tabular}{lccccccc}
\toprule
 Method       & IPO~\cite{ipo} & KTO~\cite{kto}  & StepDPO~\cite{stepdpo} & SimPO~\cite{simPO} \\ \midrule
Clean & 69.2  & 70.1 & 70.0  & 68.1\\
Adversarial & 32.6 & 32.0 & 32.3 & 30.7 \\
\bottomrule
\end{tabular}}
\caption{The performance of  DPO variants on the VQAv2 dataset.}
\label{tab:variants}
\vspace{-0.4cm}
\end{table}

\textbf{The impact of DPO variants.}
In Table~\ref{tab:variants}, we evaluate four commonly used DPO variants to analyze the effectiveness of \model. 
The results show that IPO, KTO, and StepDPO perform well, while SimPO performs relatively poorly, possibly due to the removal of the reference model.
Given the rapid advancements in DPO, further research will be necessary in the future.

\begin{table}[t]
\setlength{\abovecaptionskip}{0cm}
\centering
\setlength{\tabcolsep}{1.5mm}{
\begin{tabular}{lccccccc}
\toprule
 Method    & APGD~\cite{apgd}  & C\&W~\cite{cw}  & sC\&W~\cite{scw}  \\ \midrule
Clean & 70.0 & 70.5 & 70.3 \\
Adversarial & 32.4 & 33.1 & 32.8 \\
\bottomrule
\end{tabular}}
\caption{The evaluation of attack types on the VQAv2 dataset.}
\label{tab:attack}
\vspace{-0.5cm}
\end{table}

\textbf{Analysis of different attack types.}
We analyze the effectiveness of \model~against various types of adversarial attacks.
Table~\ref{tab:attack} illustrates that our method is insensitive to different attack methods, which may be attributed to the fact that its training samples are generated using untargeted attacks.
Additionally, our online approach exhibits strong generalization capabilities.

\begin{figure}[t]
  \centering
  \setlength{\abovecaptionskip}{0.cm}
  \includegraphics[width=\linewidth]{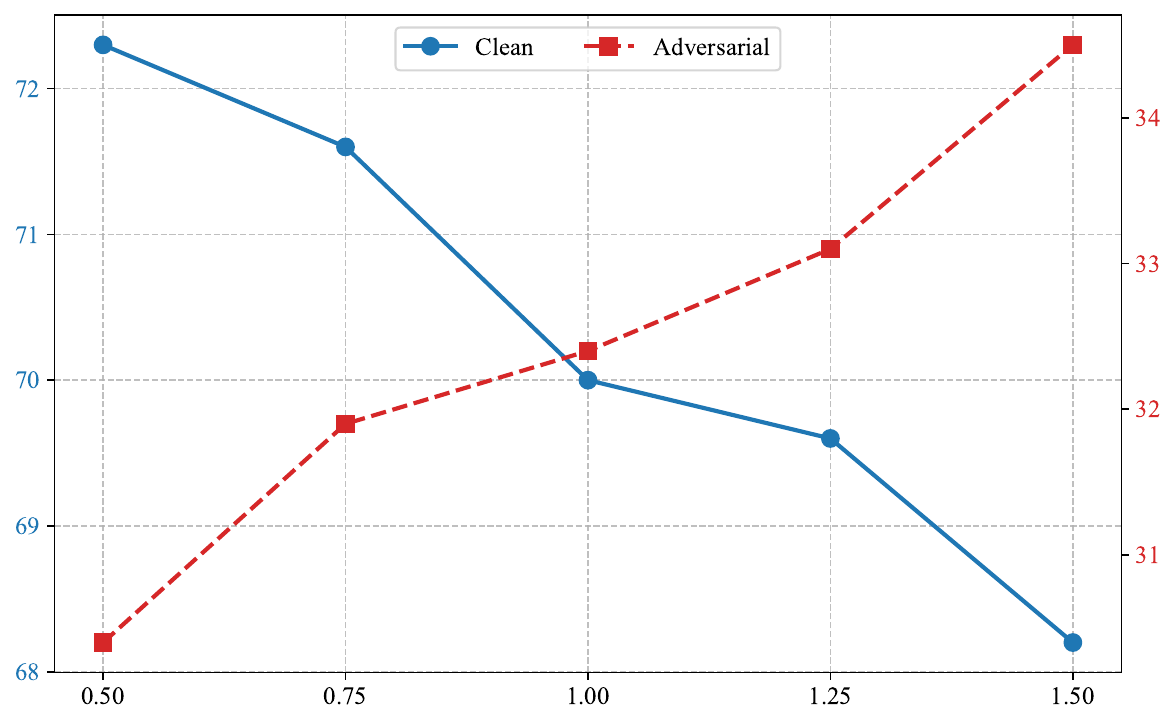 }
  \caption{Ablation study of preferred image optimization and adversarial image optimization.
  }
  \vspace{-0.7cm}
  \label{fig:abl}
\end{figure}

\textbf{The impact of $\lambda$.}
We use the setup in Section~\ref{sec:eval_untargeted_attack} to perform untargeted attacks to evaluate the effectiveness of \model~trained with diffenrent $\lambda$ on the VQAv2 dataset, with experimental results shown in Figure~\ref{fig:abl}.
Increasing $\lambda$ improves clean performance, while decreasing $\lambda$ enhances adversarial robustness. To balance adversarial and clean performance, we simply set $\lambda = 1$.


    

\section{Conclusion}
We propose \model, the first adversarial defense strategy based on preference optimization. 
The core idea of preference optimization methods, represented by DPO, is to learn both positive and negative samples simultaneously and optimize the model to better align with user preferences or goals. This is achieved by comparing the differences between positive and negative samples, clarifying the direction in which the model should be optimized.
Unlike previous adversarial fine-tuning methods, which typically only impose single-target constraints to improve adversarial robustness, leading to a loss of clean performance.
In contrast, \model~explicitly optimizes two objectives: improving adversarial robustness while maintaining proper understanding of clean images.
Both quantitative and qualitative experimental analyses demonstrate the superiority of our proposed method, offering a new perspective for future adversarial defense research.
Considering that preference optimization is gaining increasing attention in academia, introducing more refined methods into the adversarial defense field could lead to better outcomes.

{
    \small
    \bibliographystyle{ieeenat_fullname}
    \bibliography{main}

\begin{thebibliography}{119}
\providecommand{\natexlab}[1]{#1}
\providecommand{\url}[1]{\texttt{#1}}
\expandafter\ifx\csname urlstyle\endcsname\relax
  \providecommand{\doi}[1]{doi: #1}\else
  \providecommand{\doi}{doi: \begingroup \urlstyle{rm}\Url}\fi

\bibitem[Alayrac et~al.(2022)Alayrac, Donahue, Luc, Miech, Barr, Hasson, Lenc, Mensch, Millican, Reynolds, Ring, Rutherford, Cabi, Han, Gong, Samangooei, Monteiro, Menick, Borgeaud, Brock, Nematzadeh, Sharifzadeh, Binkowski, Barreira, Vinyals, Zisserman, and Simonyan]{flamingo}
Jean{-}Baptiste Alayrac, Jeff Donahue, Pauline Luc, Antoine Miech, Iain Barr, Yana Hasson, Karel Lenc, Arthur Mensch, Katherine Millican, Malcolm Reynolds, Roman Ring, Eliza Rutherford, Serkan Cabi, Tengda Han, Zhitao Gong, Sina Samangooei, Marianne Monteiro, Jacob~L. Menick, Sebastian Borgeaud, Andy Brock, Aida Nematzadeh, Sahand Sharifzadeh, Mikolaj Binkowski, Ricardo Barreira, Oriol Vinyals, Andrew Zisserman, and Kar{\'{e}}n Simonyan.
\newblock Flamingo: a visual language model for few-shot learning.
\newblock In \emph{Advances in Neural Information Processing Systems 35: Annual Conference on Neural Information Processing Systems 2022, NeurIPS 2022, New Orleans, LA, USA, November 28 - December 9, 2022}, 2022.

\bibitem[Antol et~al.(2015)Antol, Agrawal, Lu, Mitchell, Batra, Zitnick, and Parikh]{vqa}
Stanislaw Antol, Aishwarya Agrawal, Jiasen Lu, Margaret Mitchell, Dhruv Batra, C.~Lawrence Zitnick, and Devi Parikh.
\newblock {VQA:} visual question answering.
\newblock In \emph{2015 {IEEE} International Conference on Computer Vision, {ICCV} 2015, Santiago, Chile, December 7-13, 2015}, pages 2425--2433. {IEEE} Computer Society, 2015.

\bibitem[Awadalla et~al.(2023)Awadalla, Gao, Gardner, Hessel, Hanafy, Zhu, Marathe, Bitton, Gadre, Sagawa, Jitsev, Kornblith, Koh, Ilharco, Wortsman, and Schmidt]{openflamingo}
Anas Awadalla, Irena Gao, Josh Gardner, Jack Hessel, Yusuf Hanafy, Wanrong Zhu, Kalyani Marathe, Yonatan Bitton, Samir~Yitzhak Gadre, Shiori Sagawa, Jenia Jitsev, Simon Kornblith, Pang~Wei Koh, Gabriel Ilharco, Mitchell Wortsman, and Ludwig Schmidt.
\newblock Openflamingo: An open-source framework for training large autoregressive vision-language models.
\newblock \emph{CoRR}, abs/2308.01390, 2023.

\bibitem[Azar et~al.(2024)Azar, Guo, Piot, Munos, Rowland, Valko, and Calandriello]{ipo}
Mohammad~Gheshlaghi Azar, Zhaohan~Daniel Guo, Bilal Piot, R{\'{e}}mi Munos, Mark Rowland, Michal Valko, and Daniele Calandriello.
\newblock A general theoretical paradigm to understand learning from human preferences.
\newblock In \emph{International Conference on Artificial Intelligence and Statistics, 2-4 May 2024, Palau de Congressos, Valencia, Spain}, pages 4447--4455. {PMLR}, 2024.

\bibitem[Bagdasaryan et~al.(2023)Bagdasaryan, Hsieh, Nassi, and Shmatikov]{absuing}
Eugene Bagdasaryan, Tsung-Yin Hsieh, Ben Nassi, and Vitaly Shmatikov.
\newblock (ab) using images and sounds for indirect instruction injection in multi-modal llms.
\newblock \emph{arXiv preprint arXiv:2307.10490}, 2023.

\bibitem[Bai et~al.(2023)Bai, Bai, Yang, Wang, Tan, Wang, Lin, Zhou, and Zhou]{qwen-vl}
Jinze Bai, Shuai Bai, Shusheng Yang, Shijie Wang, Sinan Tan, Peng Wang, Junyang Lin, Chang Zhou, and Jingren Zhou.
\newblock Qwen-vl: A versatile vision-language model for understanding, localization, text reading, and beyond, 2023.

\bibitem[Bai et~al.(2022)Bai, Jones, Ndousse, Askell, Chen, DasSarma, Drain, Fort, Ganguli, Henighan, Joseph, Kadavath, Kernion, Conerly, Showk, Elhage, Hatfield{-}Dodds, Hernandez, Hume, Johnston, Kravec, Lovitt, Nanda, Olsson, Amodei, Brown, Clark, McCandlish, Olah, Mann, and Kaplan]{rlhf0}
Yuntao Bai, Andy Jones, Kamal Ndousse, Amanda Askell, Anna Chen, Nova DasSarma, Dawn Drain, Stanislav Fort, Deep Ganguli, Tom Henighan, Nicholas Joseph, Saurav Kadavath, Jackson Kernion, Tom Conerly, Sheer~El Showk, Nelson Elhage, Zac Hatfield{-}Dodds, Danny Hernandez, Tristan Hume, Scott Johnston, Shauna Kravec, Liane Lovitt, Neel Nanda, Catherine Olsson, Dario Amodei, Tom~B. Brown, Jack Clark, Sam McCandlish, Chris Olah, Benjamin Mann, and Jared Kaplan.
\newblock Training a helpful and harmless assistant with reinforcement learning from human feedback.
\newblock \emph{CoRR}, abs/2204.05862, 2022.

\bibitem[Bradley and Terry(1952)]{bt}
Ralph~Allan Bradley and Milton~E Terry.
\newblock Rank analysis of incomplete block designs: I. the method of paired comparisons.
\newblock \emph{Biometrika}, 39\penalty0 (3/4):\penalty0 324--345, 1952.

\bibitem[Brown et~al.(2017)Brown, Man{\'{e}}, Roy, Abadi, and Gilmer]{adv_patch}
Tom~B. Brown, Dandelion Man{\'{e}}, Aurko Roy, Mart{\'{\i}}n Abadi, and Justin Gilmer.
\newblock Adversarial patch.
\newblock \emph{CoRR}, abs/1712.09665, 2017.

\bibitem[Cao et~al.(2023{\natexlab{a}})Cao, Bao, Liu, Chen, Yin, Liu, Liu, Jiang, and Sun]{SeRum}
Haoyu Cao, Changcun Bao, Chaohu Liu, Huang Chen, Kun Yin, Hao Liu, Yinsong Liu, Deqiang Jiang, and Xing Sun.
\newblock Attention where it matters: Rethinking visual document understanding with selective region concentration.
\newblock In \emph{{IEEE/CVF} International Conference on Computer Vision, {ICCV} 2023, Paris, France, October 1-6, 2023}, pages 19460--19470. {IEEE}, 2023{\natexlab{a}}.

\bibitem[Cao et~al.(2023{\natexlab{b}})Cao, Bao, Liu, Chen, Yin, Liu, Liu, Jiang, and Sun]{cao2023attention}
Haoyu Cao, Changcun Bao, Chaohu Liu, Huang Chen, Kun Yin, Hao Liu, Yinsong Liu, Deqiang Jiang, and Xing Sun.
\newblock Attention where it matters: Rethinking visual document understanding with selective region concentration.
\newblock In \emph{Proceedings of the IEEE/CVF International Conference on Computer Vision}, pages 19517--19527, 2023{\natexlab{b}}.

\bibitem[Carlini and Wagner(2017)]{cw}
Nicholas Carlini and David Wagner.
\newblock Towards evaluating the robustness of neural networks.
\newblock In \emph{2017 ieee symposium on security and privacy (sp)}, pages 39--57. Ieee, 2017.

\bibitem[Carlini et~al.(2023)Carlini, Nasr, Choquette{-}Choo, Jagielski, Gao, Koh, Ippolito, Tram{\`{e}}r, and Schmidt]{nc2023}
Nicholas Carlini, Milad Nasr, Christopher~A. Choquette{-}Choo, Matthew Jagielski, Irena Gao, Pang~Wei Koh, Daphne Ippolito, Florian Tram{\`{e}}r, and Ludwig Schmidt.
\newblock Are aligned neural networks adversarially aligned?
\newblock In \emph{Advances in Neural Information Processing Systems 36: Annual Conference on Neural Information Processing Systems 2023, NeurIPS 2023, New Orleans, LA, USA, December 10 - 16, 2023}, 2023.

\bibitem[Chen et~al.(2024)Chen, Wang, Tian, Ye, Gao, Cui, Tong, Hu, Luo, Ma, Ma, Wang, Dong, Yan, Guo, He, Shi, Jin, Xu, Wang, Wei, Li, Zhang, Zhang, Cai, Wen, Yan, Dou, Lu, Zhu, Lu, Lin, Qiao, Dai, and Wang]{internvl2}
Zhe Chen, Weiyun Wang, Hao Tian, Shenglong Ye, Zhangwei Gao, Erfei Cui, Wenwen Tong, Kongzhi Hu, Jiapeng Luo, Zheng Ma, Ji Ma, Jiaqi Wang, Xiaoyi Dong, Hang Yan, Hewei Guo, Conghui He, Botian Shi, Zhenjiang Jin, Chao Xu, Bin Wang, Xingjian Wei, Wei Li, Wenjian Zhang, Bo Zhang, Pinlong Cai, Licheng Wen, Xiangchao Yan, Min Dou, Lewei Lu, Xizhou Zhu, Tong Lu, Dahua Lin, Yu Qiao, Jifeng Dai, and Wenhai Wang.
\newblock How far are we to gpt-4v? closing the gap to commercial multimodal models with open-source suites, 2024.

\bibitem[Chiang et~al.(2023)Chiang, Li, Lin, Sheng, Wu, Zhang, Zheng, Zhuang, Zhuang, Gonzalez, et~al.]{vicuna}
Wei-Lin Chiang, Zhuohan Li, Zi Lin, Ying Sheng, Zhanghao Wu, Hao Zhang, Lianmin Zheng, Siyuan Zhuang, Yonghao Zhuang, Joseph~E Gonzalez, et~al.
\newblock Vicuna: An open-source chatbot impressing gpt-4 with 90\%* chatgpt quality.
\newblock \emph{See https://vicuna. lmsys. org (accessed 14 April 2023)}, 2\penalty0 (3):\penalty0 6, 2023.

\bibitem[Croce and Hein(2020)]{apgd}
Francesco Croce and Matthias Hein.
\newblock Reliable evaluation of adversarial robustness with an ensemble of diverse parameter-free attacks.
\newblock In \emph{Proceedings of the 37th International Conference on Machine Learning, {ICML} 2020, 13-18 July 2020, Virtual Event}, pages 2206--2216. {PMLR}, 2020.

\bibitem[Cui et~al.(2024)Cui, Ma, Cao, Ye, Zhou, Liang, Chen, Lu, Yang, Liao, Gao, Li, Tang, Cao, Zhou, Liu, Yan, Mei, Cao, Wang, and Zheng]{survey_mllm_auto}
Can Cui, Yunsheng Ma, Xu Cao, Wenqian Ye, Yang Zhou, Kaizhao Liang, Jintai Chen, Juanwu Lu, Zichong Yang, Kuei{-}Da Liao, Tianren Gao, Erlong Li, Kun Tang, Zhipeng Cao, Tong Zhou, Ao Liu, Xinrui Yan, Shuqi Mei, Jianguo Cao, Ziran Wang, and Chao Zheng.
\newblock A survey on multimodal large language models for autonomous driving.
\newblock In \emph{{IEEE/CVF} Winter Conference on Applications of Computer Vision Workshops, {WACVW} 2024 - Workshops, Waikoloa, HI, USA, January 1-6, 2024}, pages 958--979. {IEEE}, 2024.

\bibitem[Cui et~al.(2023)Cui, Aparcedo, Jang, and Lim]{cui2024cvpr}
Xuanming Cui, Alejandro Aparcedo, Young~Kyun Jang, and Ser{-}Nam Lim.
\newblock On the robustness of large multimodal models against image adversarial attacks.
\newblock \emph{CoRR}, abs/2312.03777, 2023.

\bibitem[Dai et~al.(2023)Dai, Li, Li, Tiong, Zhao, Wang, Li, Fung, and Hoi]{instructblip}
Wenliang Dai, Junnan Li, Dongxu Li, Anthony Meng~Huat Tiong, Junqi Zhao, Weisheng Wang, Boyang Li, Pascale Fung, and Steven C.~H. Hoi.
\newblock Instructblip: Towards general-purpose vision-language models with instruction tuning.
\newblock In \emph{Advances in Neural Information Processing Systems 36: Annual Conference on Neural Information Processing Systems 2023, NeurIPS 2023, New Orleans, LA, USA, December 10 - 16, 2023}, 2023.

\bibitem[Das et~al.(2018)Das, Shanbhogue, Chen, Hohman, Li, Chen, Kounavis, and Chau]{nilaksh2018defense}
Nilaksh Das, Madhuri Shanbhogue, Shang{-}Tse Chen, Fred Hohman, Siwei Li, Li Chen, Michael~E. Kounavis, and Duen~Horng Chau.
\newblock {SHIELD:} fast, practical defense and vaccination for deep learning using {JPEG} compression.
\newblock In \emph{Proceedings of the 24th {ACM} {SIGKDD} International Conference on Knowledge Discovery {\&} Data Mining, {KDD} 2018, London, UK, August 19-23, 2018}, pages 196--204. {ACM}, 2018.

\bibitem[Debbi(2024)]{debbi2024CausAdv}
Hichem Debbi.
\newblock Causadv: A causal-based framework for detecting adversarial examples, 2024.

\bibitem[Deng et~al.(2009)Deng, Dong, Socher, Li, Li, and Fei{-}Fei]{imagenet}
Jia Deng, Wei Dong, Richard Socher, Li{-}Jia Li, Kai Li, and Li Fei{-}Fei.
\newblock Imagenet: {A} large-scale hierarchical image database.
\newblock In \emph{2009 {IEEE} Computer Society Conference on Computer Vision and Pattern Recognition {(CVPR} 2009), 20-25 June 2009, Miami, Florida, {USA}}, pages 248--255. {IEEE} Computer Society, 2009.

\bibitem[Dong et~al.(2023{\natexlab{a}})Dong, Moosavi{-}Dezfooli, Lai, and Xie]{uiat}
Junhao Dong, Seyed{-}Mohsen Moosavi{-}Dezfooli, Jianhuang Lai, and Xiaohua Xie.
\newblock The enemy of my enemy is my friend: Exploring inverse adversaries for improving adversarial training.
\newblock In \emph{{IEEE/CVF} Conference on Computer Vision and Pattern Recognition, {CVPR} 2023, Vancouver, BC, Canada, June 17-24, 2023}, pages 24678--24687. {IEEE}, 2023{\natexlab{a}}.

\bibitem[Dong et~al.(2024)Dong, Qu, Wang, and Ong]{dong2024advtrain}
Junhao Dong, Xinghua Qu, Z.~Jane Wang, and Yew-Soon Ong.
\newblock Enhancing adversarial robustness via uncertainty-aware distributional adversarial training, 2024.

\bibitem[Dong et~al.(2023{\natexlab{b}})Dong, Chen, Chen, Fang, Yang, Zhang, Tian, Su, and Zhu]{bard_robust}
Yinpeng Dong, Huanran Chen, Jiawei Chen, Zhengwei Fang, Xiao Yang, Yichi Zhang, Yu Tian, Hang Su, and Jun Zhu.
\newblock How robust is google's bard to adversarial image attacks?
\newblock \emph{CoRR}, abs/2309.11751, 2023{\natexlab{b}}.

\bibitem[Ethayarajh et~al.(2024)Ethayarajh, Xu, Muennighoff, Jurafsky, and Kiela]{kto}
Kawin Ethayarajh, Winnie Xu, Niklas Muennighoff, Dan Jurafsky, and Douwe Kiela.
\newblock Kto: Model alignment as prospect theoretic optimization.
\newblock \emph{arXiv preprint arXiv:2402.01306}, 2024.

\bibitem[Fares et~al.(2024)Fares, Ziu, Aremu, Durasov, Tak{\'{a}}c, Fua, Nandakumar, and Laptev]{MirrorCheck}
Samar Fares, Klea Ziu, Toluwani Aremu, Nikita Durasov, Martin Tak{\'{a}}c, Pascal Fua, Karthik Nandakumar, and Ivan Laptev.
\newblock Mirrorcheck: Efficient adversarial defense for vision-language models.
\newblock \emph{CoRR}, abs/2406.09250, 2024.

\bibitem[Gao et~al.(2024)Gao, Bai, Gu, Xia, Torr, Li, and Liu]{verbose_image}
Kuofeng Gao, Yang Bai, Jindong Gu, Shu{-}Tao Xia, Philip Torr, Zhifeng Li, and Wei Liu.
\newblock Inducing high energy-latency of large vision-language models with verbose images.
\newblock In \emph{The Twelfth International Conference on Learning Representations, {ICLR} 2024, Vienna, Austria, May 7-11, 2024}. OpenReview.net, 2024.

\bibitem[Goodfellow et~al.(2015)Goodfellow, Shlens, and Szegedy]{fsgm}
Ian~J. Goodfellow, Jonathon Shlens, and Christian Szegedy.
\newblock Explaining and harnessing adversarial examples.
\newblock In \emph{3rd International Conference on Learning Representations, {ICLR} 2015, San Diego, CA, USA, May 7-9, 2015, Conference Track Proceedings}, 2015.

\bibitem[Goyal et~al.(2017)Goyal, Khot, Summers{-}Stay, Batra, and Parikh]{vqav2}
Yash Goyal, Tejas Khot, Douglas Summers{-}Stay, Dhruv Batra, and Devi Parikh.
\newblock Making the {V} in {VQA} matter: Elevating the role of image understanding in visual question answering.
\newblock In \emph{2017 {IEEE} Conference on Computer Vision and Pattern Recognition, {CVPR} 2017, Honolulu, HI, USA, July 21-26, 2017}, pages 6325--6334. {IEEE} Computer Society, 2017.

\bibitem[Ho and Vasconcelos(2022)]{chin2022defense}
Chih{-}Hui Ho and Nuno Vasconcelos.
\newblock {DISCO:} adversarial defense with local implicit functions.
\newblock In \emph{Advances in Neural Information Processing Systems 35: Annual Conference on Neural Information Processing Systems 2022, NeurIPS 2022, New Orleans, LA, USA, November 28 - December 9, 2022}, 2022.

\bibitem[Hotegni and Peitz(2024)]{hotegni2024advtrain}
Sedjro~Salomon Hotegni and Sebastian Peitz.
\newblock Morel: Enhancing adversarial robustness through multi-objective representation learning, 2024.

\bibitem[Huang et~al.(2024)Huang, Zhu, Tang, Zhou, Lei, Lv, and Chua]{huang2024advdetection}
Youcheng Huang, Fengbin Zhu, Jingkun Tang, Pan Zhou, Wenqiang Lei, Jiancheng Lv, and Tat-Seng Chua.
\newblock Effective and efficient adversarial detection for vision-language models via a single vector, 2024.

\bibitem[Hurst et~al.(2024)Hurst, Lerer, Goucher, Perelman, Ramesh, Clark, Ostrow, Welihinda, Hayes, Radford, et~al.]{gpt4o}
Aaron Hurst, Adam Lerer, Adam~P Goucher, Adam Perelman, Aditya Ramesh, Aidan Clark, AJ Ostrow, Akila Welihinda, Alan Hayes, Alec Radford, et~al.
\newblock Gpt-4o system card.
\newblock \emph{arXiv preprint arXiv:2410.21276}, 2024.

\bibitem[Jia et~al.(2024{\natexlab{a}})Jia, Hu, Weng, Shi, Li, Zhang, Zhou, Liu, Luo, Huang, and Wu]{jia2024tinyllava}
Junlong Jia, Ying Hu, Xi Weng, Yiming Shi, Miao Li, Xingjian Zhang, Baichuan Zhou, Ziyu Liu, Jie Luo, Lei Huang, and Ji Wu.
\newblock Tinyllava factory: A modularized codebase for small-scale large multimodal models.
\newblock \emph{arXiv preprint arXiv:2405.11788}, 2024{\natexlab{a}}.

\bibitem[Jia et~al.(2022)Jia, Zhang, Wu, Ma, Wang, and Cao]{xiaojun2022lasat}
Xiaojun Jia, Yong Zhang, Baoyuan Wu, Ke Ma, Jue Wang, and Xiaochun Cao.
\newblock {LAS-AT:} adversarial training with learnable attack strategy.
\newblock In \emph{{IEEE/CVF} Conference on Computer Vision and Pattern Recognition, {CVPR} 2022, New Orleans, LA, USA, June 18-24, 2022}, pages 13388--13398. {IEEE}, 2022.

\bibitem[Jia et~al.(2024{\natexlab{b}})Jia, Zhang, Wei, Wu, Ma, Wang, and Cao]{xiaojun2024pgk}
Xiaojun Jia, Yong Zhang, Xingxing Wei, Baoyuan Wu, Ke Ma, Jue Wang, and Xiaochun Cao.
\newblock Improving fast adversarial training with prior-guided knowledge.
\newblock \emph{{IEEE} Trans. Pattern Anal. Mach. Intell.}, 46\penalty0 (9):\penalty0 6367--6383, 2024{\natexlab{b}}.

\bibitem[Jiang et~al.(2024)Jiang, Wang, Dong, Gui, Shi, Cao, Tang, and Kwok]{jiang2024advtrain}
Chengze Jiang, Junkai Wang, Minjing Dong, Jie Gui, Xinli Shi, Yuan Cao, Yuan~Yan Tang, and James Tin-Yau Kwok.
\newblock Improving fast adversarial training via self-knowledge guidance, 2024.

\bibitem[Kurakin et~al.(2017{\natexlab{a}})Kurakin, Goodfellow, and Bengio]{ad_ex_physicial_world}
Alexey Kurakin, Ian~J. Goodfellow, and Samy Bengio.
\newblock Adversarial examples in the physical world.
\newblock In \emph{5th International Conference on Learning Representations, {ICLR} 2017, Toulon, France, April 24-26, 2017, Workshop Track Proceedings}. OpenReview.net, 2017{\natexlab{a}}.

\bibitem[Kurakin et~al.(2017{\natexlab{b}})Kurakin, Goodfellow, and Bengio]{adml}
Alexey Kurakin, Ian~J. Goodfellow, and Samy Bengio.
\newblock Adversarial machine learning at scale.
\newblock In \emph{5th International Conference on Learning Representations, {ICLR} 2017, Toulon, France, April 24-26, 2017, Conference Track Proceedings}. OpenReview.net, 2017{\natexlab{b}}.

\bibitem[Lai et~al.(2024)Lai, Tian, Chen, Yang, Peng, and Jia]{stepdpo}
Xin Lai, Zhuotao Tian, Yukang Chen, Senqiao Yang, Xiangru Peng, and Jiaya Jia.
\newblock Step-dpo: Step-wise preference optimization for long-chain reasoning of llms.
\newblock \emph{CoRR}, abs/2406.18629, 2024.

\bibitem[Lauren{\c{c}}on et~al.(2024)Lauren{\c{c}}on, Tronchon, Cord, and Sanh]{idefics2}
Hugo Lauren{\c{c}}on, L{\'{e}}o Tronchon, Matthieu Cord, and Victor Sanh.
\newblock What matters when building vision-language models?
\newblock \emph{CoRR}, abs/2405.02246, 2024.

\bibitem[Ledda et~al.(2024)Ledda, Scodeller, Angioni, Piras, Cinà, Fumera, Biggio, and Roli]{ledda2024advtrain}
Emanuele Ledda, Giovanni Scodeller, Daniele Angioni, Giorgio Piras, Antonio~Emanuele Cinà, Giorgio Fumera, Battista Biggio, and Fabio Roli.
\newblock On the robustness of adversarial training against uncertainty attacks, 2024.

\bibitem[Li and Li(2024)]{li2024advtrain}
Binghui Li and Yuanzhi Li.
\newblock Adversarial training can provably improve robustness: Theoretical analysis of feature learning process under structured data, 2024.

\bibitem[Li et~al.(2023{\natexlab{a}})Li, Zhang, Chen, Wang, Yang, and Liu]{otter}
Bo Li, Yuanhan Zhang, Liangyu Chen, Jinghao Wang, Jingkang Yang, and Ziwei Liu.
\newblock Otter: {A} multi-modal model with in-context instruction tuning.
\newblock \emph{CoRR}, abs/2305.03726, 2023{\natexlab{a}}.

\bibitem[Li et~al.(2024)Li, Li, Wu, Tian, and Zhou]{li2024dat}
Fengpeng Li, Kemou Li, Haiwei Wu, Jinyu Tian, and Jiantao Zhou.
\newblock Dat: Improving adversarial robustness via generative amplitude mix-up in frequency domain, 2024.

\bibitem[Li et~al.(2023{\natexlab{b}})Li, Li, Savarese, and Hoi]{blip2}
Junnan Li, Dongxu Li, Silvio Savarese, and Steven C.~H. Hoi.
\newblock {BLIP-2:} bootstrapping language-image pre-training with frozen image encoders and large language models.
\newblock In \emph{International Conference on Machine Learning, {ICML} 2023, 23-29 July 2023, Honolulu, Hawaii, {USA}}, pages 19730--19742. {PMLR}, 2023{\natexlab{b}}.

\bibitem[Li et~al.(2023{\natexlab{c}})Li, Xie, Li, Chen, Wang, Chen, Yang, Wang, and Kong]{silkie}
Lei Li, Zhihui Xie, Mukai Li, Shunian Chen, Peiyi Wang, Liang Chen, Yazheng Yang, Benyou Wang, and Lingpeng Kong.
\newblock Silkie: Preference distillation for large visual language models.
\newblock \emph{CoRR}, abs/2312.10665, 2023{\natexlab{c}}.

\bibitem[Li et~al.(2023{\natexlab{d}})Li, Du, Zhou, Wang, Zhao, and Wen]{pope}
Yifan Li, Yifan Du, Kun Zhou, Jinpeng Wang, Wayne~Xin Zhao, and Ji{-}Rong Wen.
\newblock Evaluating object hallucination in large vision-language models.
\newblock In \emph{Proceedings of the 2023 Conference on Empirical Methods in Natural Language Processing, {EMNLP} 2023, Singapore, December 6-10, 2023}, pages 292--305. Association for Computational Linguistics, 2023{\natexlab{d}}.

\bibitem[Li et~al.(2023{\natexlab{e}})Li, Yang, Liu, Ma, Zhang, Yang, Sun, Liu, and Bai]{monkey}
Zhang Li, Biao Yang, Qiang Liu, Zhiyin Ma, Shuo Zhang, Jingxu Yang, Yabo Sun, Yuliang Liu, and Xiang Bai.
\newblock Monkey: Image resolution and text label are important things for large multi-modal models.
\newblock \emph{CoRR}, abs/2311.06607, 2023{\natexlab{e}}.

\bibitem[Liang et~al.(2024)Liang, Li, Niu, Shen, and Liu]{liang2024advdefensevlm}
Yuhan Liang, Yijun Li, Yumeng Niu, Qianhe Shen, and Hangyu Liu.
\newblock A hybrid defense strategy for boosting adversarial robustness in vision-language models, 2024.

\bibitem[Lin et~al.(2014)Lin, Maire, Belongie, Hays, Perona, Ramanan, Doll{\'{a}}r, and Zitnick]{coco}
Tsung{-}Yi Lin, Michael Maire, Serge~J. Belongie, James Hays, Pietro Perona, Deva Ramanan, Piotr Doll{\'{a}}r, and C.~Lawrence Zitnick.
\newblock Microsoft {COCO:} common objects in context.
\newblock In \emph{Computer Vision - {ECCV} 2014 - 13th European Conference, Zurich, Switzerland, September 6-12, 2014, Proceedings, Part {V}}, pages 740--755. Springer, 2014.

\bibitem[Liu et~al.(2024{\natexlab{a}})Liu, Yin, Cao, Jiang, Li, Liu, Jiang, Sun, and Xu]{hrvda}
Chaohu Liu, Kun Yin, Haoyu Cao, Xinghua Jiang, Xin Li, Yinsong Liu, Deqiang Jiang, Xing Sun, and Linli Xu.
\newblock {HRVDA:} high-resolution visual document assistant.
\newblock In \emph{{IEEE/CVF} Conference on Computer Vision and Pattern Recognition, {CVPR} 2024, Seattle, WA, USA, June 16-22, 2024}, pages 15534--15545. {IEEE}, 2024{\natexlab{a}}.

\bibitem[Liu et~al.(2023)Liu, Li, Wu, and Lee]{llava}
Haotian Liu, Chunyuan Li, Qingyang Wu, and Yong~Jae Lee.
\newblock Visual instruction tuning.
\newblock In \emph{Advances in Neural Information Processing Systems 36: Annual Conference on Neural Information Processing Systems 2023, NeurIPS 2023, New Orleans, LA, USA, December 10 - 16, 2023}, 2023.

\bibitem[Liu et~al.(2024{\natexlab{b}})Liu, Li, Li, and Lee]{llava1.5}
Haotian Liu, Chunyuan Li, Yuheng Li, and Yong~Jae Lee.
\newblock Improved baselines with visual instruction tuning.
\newblock In \emph{{IEEE/CVF} Conference on Computer Vision and Pattern Recognition, {CVPR} 2024, Seattle, WA, USA, June 16-22, 2024}, pages 26286--26296. {IEEE}, 2024{\natexlab{b}}.

\bibitem[Liu et~al.(2024{\natexlab{c}})Liu, Zhu, Lan, Yang, and Qiao]{survey_mllm_safety}
Xin Liu, Yichen Zhu, Yunshi Lan, Chao Yang, and Yu Qiao.
\newblock Safety of multimodal large language models on images and texts, 2024{\natexlab{c}}.

\bibitem[Liu and Chen(2024)]{trap-mid}
Zhen-Ting Liu and Shang-Tse Chen.
\newblock Trap-mid: Trapdoor-based defense against model inversion attacks, 2024.

\bibitem[Lu et~al.(2024)Lu, Liu, Zhang, Wang, Dong, Liu, Sun, Ren, Li, Yang, Sun, Deng, Xu, Xie, and Ruan]{deepseek-vl}
Haoyu Lu, Wen Liu, Bo Zhang, Bingxuan Wang, Kai Dong, Bo Liu, Jingxiang Sun, Tongzheng Ren, Zhuoshu Li, Hao Yang, Yaofeng Sun, Chengqi Deng, Hanwei Xu, Zhenda Xie, and Chong Ruan.
\newblock Deepseek-vl: Towards real-world vision-language understanding.
\newblock \emph{CoRR}, abs/2403.05525, 2024.

\bibitem[Luo et~al.(2024{\natexlab{a}})Luo, Gu, Liu, and Torr]{CroPA}
Haochen Luo, Jindong Gu, Fengyuan Liu, and Philip Torr.
\newblock An image is worth 1000 lies: Transferability of adversarial images across prompts on vision-language models.
\newblock In \emph{The Twelfth International Conference on Learning Representations, {ICLR} 2024, Vienna, Austria, May 7-11, 2024}. OpenReview.net, 2024{\natexlab{a}}.

\bibitem[Luo et~al.(2024{\natexlab{b}})Luo, Bao, Zhou, and Dang]{luo2024game}
Rui Luo, Jie Bao, Zhixin Zhou, and Chuangyin Dang.
\newblock Game-theoretic defenses for robust conformal prediction against adversarial attacks in medical imaging, 2024{\natexlab{b}}.

\bibitem[Lv et~al.(2024)Lv, Cao, Tu, Xu, Zhang, Ding, and Wang]{lv2024advtuning}
Kangtao Lv, Huangsen Cao, Kainan Tu, Yihuai Xu, Zhimeng Zhang, Xin Ding, and Yongwei Wang.
\newblock Hyper adversarial tuning for boosting adversarial robustness of pretrained large vision models, 2024.

\bibitem[Madry et~al.(2018)Madry, Makelov, Schmidt, Tsipras, and Vladu]{pgd}
Aleksander Madry, Aleksandar Makelov, Ludwig Schmidt, Dimitris Tsipras, and Adrian Vladu.
\newblock Towards deep learning models resistant to adversarial attacks.
\newblock In \emph{6th International Conference on Learning Representations, {ICLR} 2018, Vancouver, BC, Canada, April 30 - May 3, 2018, Conference Track Proceedings}. OpenReview.net, 2018.

\bibitem[Mao et~al.(2023)Mao, Geng, Yang, Wang, and Vondrick]{tecoa}
Chengzhi Mao, Scott Geng, Junfeng Yang, Xin Wang, and Carl Vondrick.
\newblock Understanding zero-shot adversarial robustness for large-scale models.
\newblock In \emph{The Eleventh International Conference on Learning Representations, {ICLR} 2023, Kigali, Rwanda, May 1-5, 2023}. OpenReview.net, 2023.

\bibitem[Mavali et~al.(2024)Mavali, Ricker, Pape, Sharma, Fischer, and Schönherr]{mavali2024detect}
Sina Mavali, Jonas Ricker, David Pape, Yash Sharma, Asja Fischer, and Lea Schönherr.
\newblock Fake it until you break it: On the adversarial robustness of ai-generated image detectors, 2024.

\bibitem[Mehta et~al.(2024)Mehta, Sekhavat, Cao, Horton, Jin, Sun, Mirzadeh, Najibi, Belenko, Zatloukal, et~al.]{Openelm}
Sachin Mehta, Mohammad~Hossein Sekhavat, Qingqing Cao, Maxwell Horton, Yanzi Jin, Chenfan Sun, Seyed~Iman Mirzadeh, Mahyar Najibi, Dmitry Belenko, Peter Zatloukal, et~al.
\newblock Openelm: An efficient language model family with open training and inference framework.
\newblock In \emph{Workshop on Efficient Systems for Foundation Models II@ ICML2024}, 2024.

\bibitem[Meng and Chen(2017)]{MagNet}
Dongyu Meng and Hao Chen.
\newblock Magnet: {A} two-pronged defense against adversarial examples.
\newblock In \emph{Proceedings of the 2017 {ACM} {SIGSAC} Conference on Computer and Communications Security, {CCS} 2017, Dallas, TX, USA, October 30 - November 03, 2017}, pages 135--147. {ACM}, 2017.

\bibitem[Meng et~al.(2025)Meng, Xia, and Chen]{simPO}
Yu Meng, Mengzhou Xia, and Danqi Chen.
\newblock Simpo: Simple preference optimization with a reference-free reward.
\newblock \emph{Advances in Neural Information Processing Systems}, 37:\penalty0 124198--124235, 2025.

\bibitem[Metzen et~al.(2017)Metzen, Genewein, Fischer, and Bischoff]{jan2017detect}
Jan~Hendrik Metzen, Tim Genewein, Volker Fischer, and Bastian Bischoff.
\newblock On detecting adversarial perturbations.
\newblock In \emph{5th International Conference on Learning Representations, {ICLR} 2017, Toulon, France, April 24-26, 2017, Conference Track Proceedings}. OpenReview.net, 2017.

\bibitem[Mumcu and Yilmaz(2024)]{mumcu2024detect}
Furkan Mumcu and Yasin Yilmaz.
\newblock Detecting adversarial examples, 2024.

\bibitem[Nguyen et~al.(2023)Nguyen, Gadre, Ilharco, Oh, and Schmidt]{improve_mm_image_captioning}
Thao Nguyen, Samir~Yitzhak Gadre, Gabriel Ilharco, Sewoong Oh, and Ludwig Schmidt.
\newblock Improving multimodal datasets with image captioning.
\newblock In \emph{Advances in Neural Information Processing Systems 36: Annual Conference on Neural Information Processing Systems 2023, NeurIPS 2023, New Orleans, LA, USA, December 10 - 16, 2023}, 2023.

\bibitem[Nie et~al.(2022)Nie, Guo, Huang, Xiao, Vahdat, and Anandkumar]{weili2022defense}
Weili Nie, Brandon Guo, Yujia Huang, Chaowei Xiao, Arash Vahdat, and Animashree Anandkumar.
\newblock Diffusion models for adversarial purification.
\newblock In \emph{International Conference on Machine Learning, {ICML} 2022, 17-23 July 2022, Baltimore, Maryland, {USA}}, pages 16805--16827. {PMLR}, 2022.

\bibitem[Niu et~al.(2024)Niu, Ren, Gao, Hua, and Jin]{jailbreaking_vlms}
Zhenxing Niu, Haodong Ren, Xinbo Gao, Gang Hua, and Rong Jin.
\newblock Jailbreaking attack against multimodal large language model.
\newblock \emph{CoRR}, abs/2402.02309, 2024.

\bibitem[Ouali et~al.(2024)Ouali, Bulat, Martinez, and Tzimiropoulos]{clipdpo}
Yassine Ouali, Adrian Bulat, Brais Martinez, and Georgios Tzimiropoulos.
\newblock Clip-dpo: Vision-language models as a source of preference for fixing hallucinations in lvlms, 2024.

\bibitem[Ouyang et~al.(2022)Ouyang, Wu, Jiang, Almeida, Wainwright, Mishkin, Zhang, Agarwal, Slama, Ray, Schulman, Hilton, Kelton, Miller, Simens, Askell, Welinder, Christiano, Leike, and Lowe]{rlhf}
Long Ouyang, Jeffrey Wu, Xu Jiang, Diogo Almeida, Carroll~L. Wainwright, Pamela Mishkin, Chong Zhang, Sandhini Agarwal, Katarina Slama, Alex Ray, John Schulman, Jacob Hilton, Fraser Kelton, Luke Miller, Maddie Simens, Amanda Askell, Peter Welinder, Paul~F. Christiano, Jan Leike, and Ryan Lowe.
\newblock Training language models to follow instructions with human feedback.
\newblock In \emph{Advances in Neural Information Processing Systems 35: Annual Conference on Neural Information Processing Systems 2022, NeurIPS 2022, New Orleans, LA, USA, November 28 - December 9, 2022}, 2022.

\bibitem[Palma et~al.(2024)Palma, Durand, Chihani, Terrier, and Urban]{depalma2024advtrain}
Alessandro~De Palma, Serge Durand, Zakaria Chihani, François Terrier, and Caterina Urban.
\newblock On using certified training towards empirical robustness, 2024.

\bibitem[Papernot et~al.(2016)Papernot, McDaniel, Wu, Jha, and Swami]{distillation_adv}
Nicolas Papernot, Patrick~D. McDaniel, Xi Wu, Somesh Jha, and Ananthram Swami.
\newblock Distillation as a defense to adversarial perturbations against deep neural networks.
\newblock In \emph{{IEEE} Symposium on Security and Privacy, {SP} 2016, San Jose, CA, USA, May 22-26, 2016}, pages 582--597. {IEEE} Computer Society, 2016.

\bibitem[Plummer et~al.(2015)Plummer, Wang, Cervantes, Caicedo, Hockenmaier, and Lazebnik]{flickr30k}
Bryan~A. Plummer, Liwei Wang, Chris~M. Cervantes, Juan~C. Caicedo, Julia Hockenmaier, and Svetlana Lazebnik.
\newblock Flickr30k entities: Collecting region-to-phrase correspondences for richer image-to-sentence models.
\newblock In \emph{2015 {IEEE} International Conference on Computer Vision, {ICCV} 2015, Santiago, Chile, December 7-13, 2015}, pages 2641--2649. {IEEE} Computer Society, 2015.

\bibitem[Qi et~al.(2024{\natexlab{a}})Qi, Huang, Panda, Henderson, Wang, and Mittal]{qi2024aaai}
Xiangyu Qi, Kaixuan Huang, Ashwinee Panda, Peter Henderson, Mengdi Wang, and Prateek Mittal.
\newblock Visual adversarial examples jailbreak aligned large language models.
\newblock In \emph{Thirty-Eighth {AAAI} Conference on Artificial Intelligence, {AAAI} 2024, Thirty-Sixth Conference on Innovative Applications of Artificial Intelligence, {IAAI} 2024, Fourteenth Symposium on Educational Advances in Artificial Intelligence, {EAAI} 2014, February 20-27, 2024, Vancouver, Canada}, pages 21527--21536. {AAAI} Press, 2024{\natexlab{a}}.

\bibitem[Qi et~al.(2024{\natexlab{b}})Qi, Huang, Panda, Henderson, Wang, and Mittal]{visual_jailbreak}
Xiangyu Qi, Kaixuan Huang, Ashwinee Panda, Peter Henderson, Mengdi Wang, and Prateek Mittal.
\newblock Visual adversarial examples jailbreak aligned large language models.
\newblock In \emph{Thirty-Eighth {AAAI} Conference on Artificial Intelligence, {AAAI} 2024, Thirty-Sixth Conference on Innovative Applications of Artificial Intelligence, {IAAI} 2024, Fourteenth Symposium on Educational Advances in Artificial Intelligence, {EAAI} 2014, February 20-27, 2024, Vancouver, Canada}, pages 21527--21536. {AAAI} Press, 2024{\natexlab{b}}.

\bibitem[Radford et~al.(2021)Radford, Kim, Hallacy, Ramesh, Goh, Agarwal, Sastry, Askell, Mishkin, Clark, Krueger, and Sutskever]{clip}
Alec Radford, Jong~Wook Kim, Chris Hallacy, Aditya Ramesh, Gabriel Goh, Sandhini Agarwal, Girish Sastry, Amanda Askell, Pamela Mishkin, Jack Clark, Gretchen Krueger, and Ilya Sutskever.
\newblock Learning transferable visual models from natural language supervision.
\newblock In \emph{Proceedings of the 38th International Conference on Machine Learning, {ICML} 2021, 18-24 July 2021, Virtual Event}, pages 8748--8763. {PMLR}, 2021.

\bibitem[Rafailov et~al.(2023)Rafailov, Sharma, Mitchell, Manning, Ermon, and Finn]{dpo}
Rafael Rafailov, Archit Sharma, Eric Mitchell, Christopher~D. Manning, Stefano Ermon, and Chelsea Finn.
\newblock Direct preference optimization: Your language model is secretly a reward model.
\newblock In \emph{Advances in Neural Information Processing Systems 36: Annual Conference on Neural Information Processing Systems 2023, NeurIPS 2023, New Orleans, LA, USA, December 10 - 16, 2023}, 2023.

\bibitem[RIbeiro et~al.(2024)RIbeiro, Schön, Zahariah, and Bach]{ribeiro2024advtrain}
Antônio~H. RIbeiro, Thomas~B. Schön, Dave Zahariah, and Francis Bach.
\newblock Efficient optimization algorithms for linear adversarial training, 2024.

\bibitem[Roth et~al.(2019)Roth, Kilcher, and Hofmann]{kevin2019detect}
Kevin Roth, Yannic Kilcher, and Thomas Hofmann.
\newblock The odds are odd: {A} statistical test for detecting adversarial examples.
\newblock In \emph{Proceedings of the 36th International Conference on Machine Learning, {ICML} 2019, 9-15 June 2019, Long Beach, California, {USA}}, pages 5498--5507. {PMLR}, 2019.

\bibitem[Samangouei et~al.(2018)Samangouei, Kabkab, and Chellappa]{pouya2018defense}
Pouya Samangouei, Maya Kabkab, and Rama Chellappa.
\newblock Defense-gan: Protecting classifiers against adversarial attacks using generative models.
\newblock In \emph{6th International Conference on Learning Representations, {ICLR} 2018, Vancouver, BC, Canada, April 30 - May 3, 2018, Conference Track Proceedings}. OpenReview.net, 2018.

\bibitem[Schlarmann and Hein(2023)]{chiccv2023}
Christian Schlarmann and Matthias Hein.
\newblock On the adversarial robustness of multi-modal foundation models.
\newblock In \emph{{IEEE/CVF} International Conference on Computer Vision, {ICCV} 2023 - Workshops, Paris, France, October 2-6, 2023}, pages 3679--3687. {IEEE}, 2023.

\bibitem[Schlarmann et~al.(2024)Schlarmann, Singh, Croce, and Hein]{robustclip}
Christian Schlarmann, Naman~Deep Singh, Francesco Croce, and Matthias Hein.
\newblock Robust {CLIP:} unsupervised adversarial fine-tuning of vision embeddings for robust large vision-language models.
\newblock In \emph{Forty-first International Conference on Machine Learning, {ICML} 2024, Vienna, Austria, July 21-27, 2024}. OpenReview.net, 2024.

\bibitem[Shayegani et~al.(2024)Shayegani, Dong, and Abu{-}Ghazaleh]{jailbreak_pieces}
Erfan Shayegani, Yue Dong, and Nael~B. Abu{-}Ghazaleh.
\newblock Jailbreak in pieces: Compositional adversarial attacks on multi-modal language models.
\newblock In \emph{The Twelfth International Conference on Learning Representations, {ICLR} 2024, Vienna, Austria, May 7-11, 2024}. OpenReview.net, 2024.

\bibitem[Singh et~al.(2019)Singh, Natarajan, Shah, Jiang, Chen, Batra, Parikh, and Rohrbach]{textvqa}
Amanpreet Singh, Vivek Natarajan, Meet Shah, Yu Jiang, Xinlei Chen, Dhruv Batra, Devi Parikh, and Marcus Rohrbach.
\newblock Towards {VQA} models that can read.
\newblock In \emph{{IEEE} Conference on Computer Vision and Pattern Recognition, {CVPR} 2019, Long Beach, CA, USA, June 16-20, 2019}, pages 8317--8326. Computer Vision Foundation / {IEEE}, 2019.

\bibitem[Szegedy et~al.(2014)Szegedy, Zaremba, Sutskever, Bruna, Erhan, Goodfellow, and Fergus]{adv_example}
Christian Szegedy, Wojciech Zaremba, Ilya Sutskever, Joan Bruna, Dumitru Erhan, Ian~J. Goodfellow, and Rob Fergus.
\newblock Intriguing properties of neural networks.
\newblock In \emph{2nd International Conference on Learning Representations, {ICLR} 2014, Banff, AB, Canada, April 14-16, 2014, Conference Track Proceedings}, 2014.

\bibitem[Team et~al.(2023)]{mpt7b}
MosaicML~NLP Team et~al.
\newblock Introducing mpt-7b: A new standard for open-source, commercially usable llms, 2023.

\bibitem[Tram{\`{e}}r et~al.(2018)Tram{\`{e}}r, Kurakin, Papernot, Goodfellow, Boneh, and McDaniel]{ensembleadvtrain}
Florian Tram{\`{e}}r, Alexey Kurakin, Nicolas Papernot, Ian~J. Goodfellow, Dan Boneh, and Patrick~D. McDaniel.
\newblock Ensemble adversarial training: Attacks and defenses.
\newblock In \emph{6th International Conference on Learning Representations, {ICLR} 2018, Vancouver, BC, Canada, April 30 - May 3, 2018, Conference Track Proceedings}. OpenReview.net, 2018.

\bibitem[Vedantam et~al.(2015)Vedantam, Zitnick, and Parikh]{cider}
Ramakrishna Vedantam, C.~Lawrence Zitnick, and Devi Parikh.
\newblock Cider: Consensus-based image description evaluation.
\newblock In \emph{{IEEE} Conference on Computer Vision and Pattern Recognition, {CVPR} 2015, Boston, MA, USA, June 7-12, 2015}, pages 4566--4575. {IEEE} Computer Society, 2015.

\bibitem[Wang et~al.(2024{\natexlab{a}})Wang, Zhou, Huang, Xu, Zhang, Poon, and Chen]{mdpo}
Fei Wang, Wenxuan Zhou, James~Y. Huang, Nan Xu, Sheng Zhang, Hoifung Poon, and Muhao Chen.
\newblock mdpo: Conditional preference optimization for multimodal large language models.
\newblock \emph{CoRR}, abs/2406.11839, 2024{\natexlab{a}}.

\bibitem[Wang et~al.(2024{\natexlab{b}})Wang, Long, Fan, and Wei]{from_llm_to_mllm_jailbreak}
Siyuan Wang, Zhuohan Long, Zhihao Fan, and Zhongyu Wei.
\newblock From llms to mllms: Exploring the landscape of multimodal jailbreaking.
\newblock \emph{CoRR}, abs/2406.14859, 2024{\natexlab{b}}.

\bibitem[Wang et~al.(2024{\natexlab{c}})Wang, Zhang, Yuan, and Shan]{PMG-AFT}
Sibo Wang, Jie Zhang, Zheng Yuan, and Shiguang Shan.
\newblock Pre-trained model guided fine-tuning for zero-shot adversarial robustness.
\newblock In \emph{Proceedings of the IEEE/CVF conference on computer vision and pattern recognition}, pages 24502--24511, 2024{\natexlab{c}}.

\bibitem[Wang et~al.(2024{\natexlab{d}})Wang, Chen, Wang, Zhou, Zhou, Yao, Zhou, Goldstein, Bhatia, Huang, and Xiao]{sima}
Xiyao Wang, Jiuhai Chen, Zhaoyang Wang, Yuhang Zhou, Yiyang Zhou, Huaxiu Yao, Tianyi Zhou, Tom Goldstein, Parminder Bhatia, Furong Huang, and Cao Xiao.
\newblock Enhancing visual-language modality alignment in large vision language models via self-improvement.
\newblock \emph{CoRR}, abs/2405.15973, 2024{\natexlab{d}}.

\bibitem[Wang et~al.(2024{\natexlab{e}})Wang, Liu, yanqiuqu, Cao, Jiang, and Xu]{wang2024break}
Yubo Wang, Chaohu Liu, yanqiuqu, Haoyu Cao, Deqiang Jiang, and Linli Xu.
\newblock Break the visual perception: Adversarial attacks targeting encoded visual tokens of large vision-language models.
\newblock In \emph{ACM Multimedia 2024}, 2024{\natexlab{e}}.

\bibitem[Wang et~al.(2025)Wang, Tang, Liu, and Xu]{wang2025tracking}
Yubo Wang, Jianting Tang, Chaohu Liu, and Linli Xu.
\newblock Tracking the copyright of large vision-language models through parameter learning adversarial images.
\newblock \emph{arXiv preprint arXiv:2502.16593}, 2025.

\bibitem[Wang et~al.(2024{\natexlab{f}})Wang, Bi, Pentyala, Ramnath, Chaudhuri, Mehrotra, Zhu, Mao, Asur, and Cheng]{dpo_survey}
Zhichao Wang, Bin Bi, Shiva~Kumar Pentyala, Kiran Ramnath, Sougata Chaudhuri, Shubham Mehrotra, Zixu Zhu, Xiang{-}Bo Mao, Sitaram Asur, and Na Cheng.
\newblock A comprehensive survey of {LLM} alignment techniques: Rlhf, rlaif, ppo, {DPO} and more.
\newblock \emph{CoRR}, abs/2407.16216, 2024{\natexlab{f}}.

\bibitem[Wang et~al.(2024{\natexlab{g}})Wang, Li, Zhu, and Xie]{AdvXL}
Zeyu Wang, Xianhang Li, Hongru Zhu, and Cihang Xie.
\newblock Revisiting adversarial training at scale.
\newblock In \emph{Proceedings of the IEEE/CVF Conference on Computer Vision and Pattern Recognition}, pages 24675--24685, 2024{\natexlab{g}}.

\bibitem[Xu et~al.(2018)Xu, Evans, and Qi]{weilin2018detect}
Weilin Xu, David Evans, and Yanjun Qi.
\newblock Feature squeezing: Detecting adversarial examples in deep neural networks.
\newblock In \emph{25th Annual Network and Distributed System Security Symposium, {NDSS} 2018, San Diego, California, USA, February 18-21, 2018}. The Internet Society, 2018.

\bibitem[Xue et~al.(2024)Xue, Wang, Qin, and Pedarsani]{xue2024advtrain}
Zhiyu Xue, Haohan Wang, Yao Qin, and Ramtin Pedarsani.
\newblock Conflict-aware adversarial training, 2024.

\bibitem[Yao et~al.(2024)Yao, Yu, Zhang, Wang, Cui, Zhu, Cai, Li, Zhao, He, Chen, Zhou, Zou, Zhang, Hu, Zheng, Zhou, Cai, Han, Zeng, Li, Liu, and Sun]{minicpm-v}
Yuan Yao, Tianyu Yu, Ao Zhang, Chongyi Wang, Junbo Cui, Hongji Zhu, Tianchi Cai, Haoyu Li, Weilin Zhao, Zhihui He, Qianyu Chen, Huarong Zhou, Zhensheng Zou, Haoye Zhang, Shengding Hu, Zhi Zheng, Jie Zhou, Jie Cai, Xu Han, Guoyang Zeng, Dahai Li, Zhiyuan Liu, and Maosong Sun.
\newblock Minicpm-v: A gpt-4v level mllm on your phone, 2024.

\bibitem[Ye et~al.(2023)Ye, Xu, Xu, Ye, Yan, Zhou, Wang, Hu, Shi, Shi, Li, Xu, Chen, Tian, Qi, Zhang, and Huang]{mPLUG-Owl}
Qinghao Ye, Haiyang Xu, Guohai Xu, Jiabo Ye, Ming Yan, Yiyang Zhou, Junyang Wang, Anwen Hu, Pengcheng Shi, Yaya Shi, Chenliang Li, Yuanhong Xu, Hehong Chen, Junfeng Tian, Qian Qi, Ji Zhang, and Fei Huang.
\newblock mplug-owl: Modularization empowers large language models with multimodality.
\newblock \emph{CoRR}, abs/2304.14178, 2023.

\bibitem[Yin et~al.(2023)Yin, Fu, Zhao, Li, Sun, Xu, and Chen]{mllmsurvey2023}
Shukang Yin, Chaoyou Fu, Sirui Zhao, Ke Li, Xing Sun, Tong Xu, and Enhong Chen.
\newblock A survey on multimodal large language models.
\newblock \emph{CoRR}, abs/2306.13549, 2023.

\bibitem[Yu et~al.(2025)Yu, Zhang, and Xu]{TGA-ZSR}
Lu Yu, Haiyang Zhang, and Changsheng Xu.
\newblock Text-guided attention is all you need for zero-shot robustness in vision-language models.
\newblock \emph{Advances in Neural Information Processing Systems}, 37:\penalty0 96424--96448, 2025.

\bibitem[Yu et~al.(2023)Yu, Yao, Zhang, He, Han, Cui, Hu, Liu, Zheng, Sun, and Chua]{rlhfv}
Tianyu Yu, Yuan Yao, Haoye Zhang, Taiwen He, Yifeng Han, Ganqu Cui, Jinyi Hu, Zhiyuan Liu, Hai{-}Tao Zheng, Maosong Sun, and Tat{-}Seng Chua.
\newblock {RLHF-V:} towards trustworthy mllms via behavior alignment from fine-grained correctional human feedback.
\newblock \emph{CoRR}, abs/2312.00849, 2023.

\bibitem[Yu et~al.(2024)Yu, Zhang, Yao, Dang, Chen, Lu, Cui, He, Liu, Chua, and Sun]{rlaifv}
Tianyu Yu, Haoye Zhang, Yuan Yao, Yunkai Dang, Da Chen, Xiaoman Lu, Ganqu Cui, Taiwen He, Zhiyuan Liu, Tat{-}Seng Chua, and Maosong Sun.
\newblock {RLAIF-V:} aligning mllms through open-source {AI} feedback for super {GPT-4V} trustworthiness.
\newblock \emph{CoRR}, abs/2405.17220, 2024.

\bibitem[Zhai et~al.(2023)Zhai, Mustafa, Kolesnikov, and Beyer]{sigclip}
Xiaohua Zhai, Basil Mustafa, Alexander Kolesnikov, and Lucas Beyer.
\newblock Sigmoid loss for language image pre-training.
\newblock In \emph{{IEEE/CVF} International Conference on Computer Vision, {ICCV} 2023, Paris, France, October 1-6, 2023}, pages 11941--11952. {IEEE}, 2023.

\bibitem[Zhang et~al.(2024)Zhang, Xu, Wu, Liu, and Zhou]{zhang2024adv10year}
Chiyu Zhang, Xiaogang Xu, Jiafei Wu, Zhe Liu, and Lu Zhou.
\newblock Adversarial attacks of vision tasks in the past 10 years: A survey, 2024.

\bibitem[Zhang et~al.(2020)Zhang, Avrithis, Furon, and Amsaleg]{scw}
Hanwei Zhang, Yannis Avrithis, Teddy Furon, and Laurent Amsaleg.
\newblock Smooth adversarial examples.
\newblock \emph{EURASIP Journal on Information Security}, 2020:\penalty0 1--12, 2020.

\bibitem[Zhang et~al.(2023)Zhang, Huang, Wu, and Lyu]{tgr}
Jianping Zhang, Yizhan Huang, Weibin Wu, and Michael~R. Lyu.
\newblock Transferable adversarial attacks on vision transformers with token gradient regularization.
\newblock In \emph{{IEEE/CVF} Conference on Computer Vision and Pattern Recognition, {CVPR} 2023, Vancouver, BC, Canada, June 17-24, 2023}, pages 16415--16424. {IEEE}, 2023.

\bibitem[Zhao et~al.(2024)Zhao, Zhang, Ye, Lu, Yin, and Wang]{zhao2024advtrainsurvey}
Mengnan Zhao, Lihe Zhang, Jingwen Ye, Huchuan Lu, Baocai Yin, and Xinchao Wang.
\newblock Adversarial training: A survey, 2024.

\bibitem[Zhao et~al.(2023{\natexlab{a}})Zhao, Zhou, Li, Tang, Wang, Hou, Min, Zhang, Zhang, Dong, Du, Yang, Chen, Chen, Jiang, Ren, Li, Tang, Liu, Liu, Nie, and Wen]{survey_llm}
Wayne~Xin Zhao, Kun Zhou, Junyi Li, Tianyi Tang, Xiaolei Wang, Yupeng Hou, Yingqian Min, Beichen Zhang, Junjie Zhang, Zican Dong, Yifan Du, Chen Yang, Yushuo Chen, Zhipeng Chen, Jinhao Jiang, Ruiyang Ren, Yifan Li, Xinyu Tang, Zikang Liu, Peiyu Liu, Jian{-}Yun Nie, and Ji{-}Rong Wen.
\newblock A survey of large language models.
\newblock \emph{CoRR}, abs/2303.18223, 2023{\natexlab{a}}.

\bibitem[Zhao et~al.(2023{\natexlab{b}})Zhao, Pang, Du, Yang, Li, Cheung, and Lin]{zhao2023nips}
Yunqing Zhao, Tianyu Pang, Chao Du, Xiao Yang, Chongxuan Li, Ngai{-}Man Cheung, and Min Lin.
\newblock On evaluating adversarial robustness of large vision-language models.
\newblock In \emph{Advances in Neural Information Processing Systems 36: Annual Conference on Neural Information Processing Systems 2023, NeurIPS 2023, New Orleans, LA, USA, December 10 - 16, 2023}, 2023{\natexlab{b}}.

\bibitem[Zhou et~al.(2024)Zhou, Hu, Weng, Jia, Luo, Liu, Wu, and Huang]{TinyLLaVA}
Baichuan Zhou, Ying Hu, Xi Weng, Junlong Jia, Jie Luo, Xien Liu, Ji Wu, and Lei Huang.
\newblock Tinyllava: A framework of small-scale large multimodal models, 2024.

\bibitem[Zhou and Patel(2022)]{deep_metric_learning}
Mo Zhou and Vishal~M. Patel.
\newblock Enhancing adversarial robustness for deep metric learning.
\newblock In \emph{{IEEE/CVF} Conference on Computer Vision and Pattern Recognition, {CVPR} 2022, New Orleans, LA, USA, June 18-24, 2022}, pages 15304--15313. {IEEE}, 2022.

\bibitem[Zhou et~al.(2023)Zhou, Hu, Li, Zhang, Zhang, and Jin]{advclip}
Ziqi Zhou, Shengshan Hu, Minghui Li, Hangtao Zhang, Yechao Zhang, and Hai Jin.
\newblock Advclip: Downstream-agnostic adversarial examples in multimodal contrastive learning.
\newblock In \emph{Proceedings of the 31st {ACM} International Conference on Multimedia, {MM} 2023, Ottawa, ON, Canada, 29 October 2023- 3 November 2023}, pages 6311--6320. {ACM}, 2023.

\bibitem[Zhu et~al.(2024)Zhu, Chen, Shen, Li, and Elhoseiny]{minigpt4}
Deyao Zhu, Jun Chen, Xiaoqian Shen, Xiang Li, and Mohamed Elhoseiny.
\newblock Minigpt-4: Enhancing vision-language understanding with advanced large language models.
\newblock In \emph{The Twelfth International Conference on Learning Representations, {ICLR} 2024, Vienna, Austria, May 7-11, 2024}. OpenReview.net, 2024.

\end{thebibliography}
}

\end{document}